\documentclass[10pt,twocolumn,letterpaper]{article}

\usepackage[pagenumbers]{cvpr}      

\usepackage[T1]{fontenc}
\usepackage{cuted} % CVPR 2027 two-column teaser
\providecommand{\methodname}{HOForce\xspace}
\newsavebox{\CompactExperimentBox}
\definecolor{cvprblue}{rgb}{0.21,0.49,0.74}
\usepackage[breaklinks,colorlinks,allcolors=cvprblue]{hyperref}

\def\confName{CVPR}
\def\confYear{2027}

\title{HOForce: Motion and Force Reconstruction of Hand--Object Interaction\\Using a Single-View RGB-D Video}

\author{
Jiahui Wang$^{1,2}$\thanks{Equal contribution.}\quad Wanqing Du$^{1,2,*}$\quad Yuanbin Zou$^{2}$\quad
Zhefeng Xu$^{3}$\quad Feng Xu$^{1}$\thanks{Corresponding author.}\\[4pt]
$^{1}$Tsinghua University\qquad $^{2}$Parity Lab\\
$^{3}$Hangzhou Dianzi University
}

\hypersetup{
  pdftitle={HOForce: Motion and Force Reconstruction of Hand-Object Interaction Using a Single-View RGB-D Video},
  pdfauthor={Jiahui Wang, Wanqing Du, Yuanbin Zou, Zhefeng Xu, Feng Xu}
}
\begin{document}
\maketitle
\begingroup
\setlength{\stripsep}{5pt plus 1pt minus 1pt}
\begin{strip}
\centering
\includegraphics[width=0.90\linewidth]{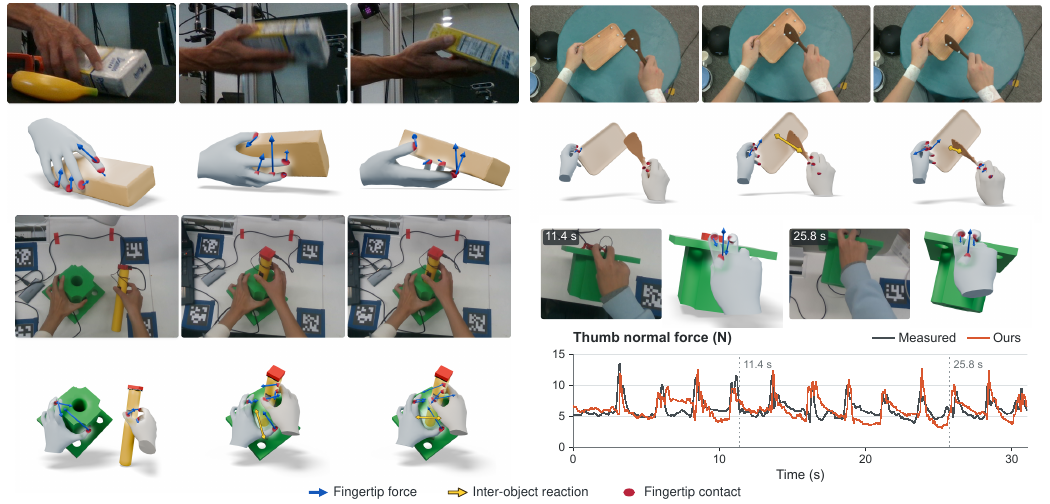}
\captionsetup{hypcap=false}
\captionof{figure}{\textbf{\methodname reconstructs hand--object motion and contact forces from single-view RGB-D video.} Examples span DexYCB, TACO, and captured sequences. Arrows show fingertip forces and simulated inter-object reactions at separate scales. Red regions mark fingertip contacts.}
\captionsetup{hypcap=true}
\label{fig:reconstruction_teaser}%
\end{strip}
\endgroup

\begin{abstract}
Reconstructing hand--object interactions is an important task, and achieving it with a single camera makes large-scale capture practical but challenging due to heavy occlusions.
We introduce \methodname to track hands and objects, and further recover interaction forces from single-view RGB-D video and known object properties.
For two-hand manipulation, we jointly model fingertip forces of both hands and the contact forces between objects when two objects are presented.
The reconstructed forces also improve motion tracking by giving pose constraints in the occluded regions.
\methodname recovers hand and object motion more accurately on DexYCB and TACO and produces less hand--object penetration than interaction baselines on DexYCB.
On our dataset of fingertip forces recorded with tactile sensors, \methodname reduces normal-force MAE by 24.5\% relative to the strongest baseline.
The recovered forces reproduce object motion in MuJoCo and transfer to other engines after adaptation.
See our \href{https://huggingface.co/spaces/fnoi16wjhui/HOForce}{project page}.
\end{abstract}

\section{Introduction}
\label{sec:introduction}
Reconstructing motion and contact forces of hand--object interactions is an important task in computer vision, graphics, and embodied AI. 
It reveals how people grasp and manipulate objects, providing motion for character animation and demonstrations for robot learning, and the contact forces explain how the fingers support, turn, or push an object to produce that motion.
Achieving this using a single RGB-D camera makes it practical to record at scale in everyday settings without installing and calibrating a multi-camera rig. 
%Most visual reconstructions describe how the hands and objects move~\cite{chen2023hotrack,shi2026agile}. 

A single view leaves an interaction underdetermined in two major aspects. 
First, large occlusions caused by single-view capture involve strong ambiguities such that several hand poses can explain the same visible surface while placing the fingertips at different contacts. Second, the camera does not measure the forces at those contacts. 
Tactile sensors provide direct measurements but require instrumented capture. 
Visual pressure predictors~\cite{zhang2026egotac,jeon2026hope} estimate normal loads, leaving the tangential forces that resist slip unresolved. 
Physics-based methods use simulated motion to refine poses~\cite{hu2024hoic} or infer fingertip forces from object dynamics~\cite{hu2022physicalinteraction}, but when two hands manipulate two objects in contact, they cannot jointly recover the forces exerted by the fingertips and those transmitted between the objects. 
Fitting each object independently can therefore misattribute inter-object forces to the fingers.

We introduce \methodname, an inverse dynamics method for reconstructing hand--object interactions. 
Given single-view RGB-D video and known object shape, weight, and friction, it uses the observed motion to infer where the fingers make contact and what forces they apply. 
When objects touch, the solver also accounts for the forces they exert on each other. 
The estimated forces in turn help refine the hand pose.

To recover the forces, we simulate the reconstructed scene in MuJoCo~\cite{todorov2012mujoco}, which resolves inter-object and support reactions. 
We fit fingertip forces to the reconstructed trajectories by locally linearizing the motion error and evaluating each update in simulation. 
With these forces fixed, we refine contact locations and poses, penalizing gaps more strongly at contacts that carry larger loads. 
Alternating these updates yields forces that can be reapplied to reproduce the manipulation.

This inverse problem requires an accurate motion estimate to anchor the solution to the video. 
We initialize hand and object poses from RGB-D observations and instance masks, compensate for fast object motion, and re-register objects after tracking loss. 
We then jointly refine hand and object trajectories across frames. 
Distance-based penetration penalties~\cite{cao2021rho,aytekin2026grag} can move a finger toward the wrong face of a thin wall. 
We use the finger bone axis to identify the contact side and guide the penetration correction.

Overall, our main contributions are threefold:
\begin{itemize}
  \item We propose a single-view RGB-D framework that recovers hand and object motion together with interaction forces, including bimanual manipulation with load transfer between objects.
  \item We couple force estimation with contact and pose refinement, using the dynamics of interacting objects to constrain fingertip forces and improve visually ambiguous grasps.
  \item We demonstrate improved motion reconstruction and interaction geometry on DexYCB and TACO. We validate the recovered fingertip forces against tactile measurements and by replaying object motion in simulation.
\end{itemize}

\section{Related Work}
\label{sec:related_work}

\subsection{Visual Hand-Object Reconstruction}
\label{sec:related_visual}
\paragraph{Object pose estimation and tracking.}
RGB-D methods estimate rigid poses by combining appearance and geometry~\cite{wang2019densefusion,stoiber2022icg}. FoundationPose~\cite{wen2024foundationpose} generalizes pose estimation and tracking to novel objects given a CAD model or reference images, using rendered observations for learned pose refinement and hypothesis scoring. We initialize object motion with its model-based tracker. Beyond known meshes, BundleSDF~\cite{wen2023bundlesdf} jointly recovers shape and motion, while Point2Pose~\cite{lin2026point2pose} uses long-range point correspondences for tracking and recovery after occlusion.

\paragraph{Hand reconstruction.}
Image-based methods recover articulation and shape through parametric models or direct mesh regression~\cite{romero2017mano,boukhayma2019hand,lin2021metro}. HaMeR~\cite{pavlakos2024hamer} improves mesh estimation through a large transformer trained on diverse hand images. WiLoR~\cite{potamias2025wilor} integrates hand localization with MANO reconstruction and refines mesh alignment using local image features. WiLoR supplies our initial hand articulation. HaWoR and Dyn-HaMR further recover global hand motion by accounting for camera movement and temporal continuity~\cite{zhang2025hawor,yu2025dynhamr}.

\paragraph{Joint hand--object reconstruction.}
Early methods jointly fit hands and objects using geometric alignment and interaction constraints in RGB-D~\cite{tzionas2016capturing,sridhar2016realtimeho,tsoli2018joint,hampali2020honnotate,zhang2021singledepth} or RGB video~\cite{hasson2021unconstrained}. HOTrack~\cite{chen2023hotrack} combines point-cloud hand tracking with SDF-based object tracking and interaction refinement. Learning-based approaches exploit joint shape prediction, contact distributions, or implicit geometric cues~\cite{hasson2019joint,grady2021contactopt,qi2024hoisdf}. More recently, HOLD and EasyHOI reconstruct unknown object geometry together with hand--object configurations~\cite{fan2024hold,liu2025easyhoi}, while AGILE~\cite{shi2026agile} combines generative shape priors, predicted depth, and motion optimization. Given metric meshes and measured depth, we focus on recovering camera-space hand and object trajectories while maintaining consistent grasps at thin surfaces.

\subsection{Physics-Based Interaction Reconstruction and Contact Force Estimation}
\label{sec:related_physics}
Prior work uses object dynamics to infer contact forces from observed motion~\cite{ehsani2020force}.
Pham et al.~\cite{pham2018force} combine learned predictions with
second-order cone programming to estimate finger forces.
Li et al.~\cite{li2019motionforces} use trajectory optimization to jointly
recover human and object motion, contact locations, and forces.
Physical Interaction~\cite{hu2022physicalinteraction} estimates fingertip
forces and contact states from object dynamics and uses kinematic
confidence to regulate contact sliding, improving hand--object interaction
reconstruction.

Recent methods predict surface contact and pressure distributions from
visual observations. EgoTac~\cite{zhang2026egotac} estimates hand-surface
contact and normal pressure from egocentric video.
HOPE~\cite{jeon2026hope} combines video features with hand poses to recover
time-varying contact and pressure.
EgoPHI~\cite{ilic2026egophi} uses physics simulation to generate force
supervision and predicts contact and normal force distributions on both
hands and object surfaces from a single image and object geometry.

Our second stage combines visual confidence, contact geometry, and hand
kinematics to jointly optimize contacts, poses, and hand--object forces for
bimanual multi-object interaction in one dynamics system. The simulator
resolves object--object and support reactions, allowing object motion and
interobject load transfer to jointly constrain force estimation.

\section{Method}
\label{sec:method}

\subsection{Overview}
\label{sec:overview}
\begin{figure*}[t]
  \centering
  \includegraphics[width=\textwidth]{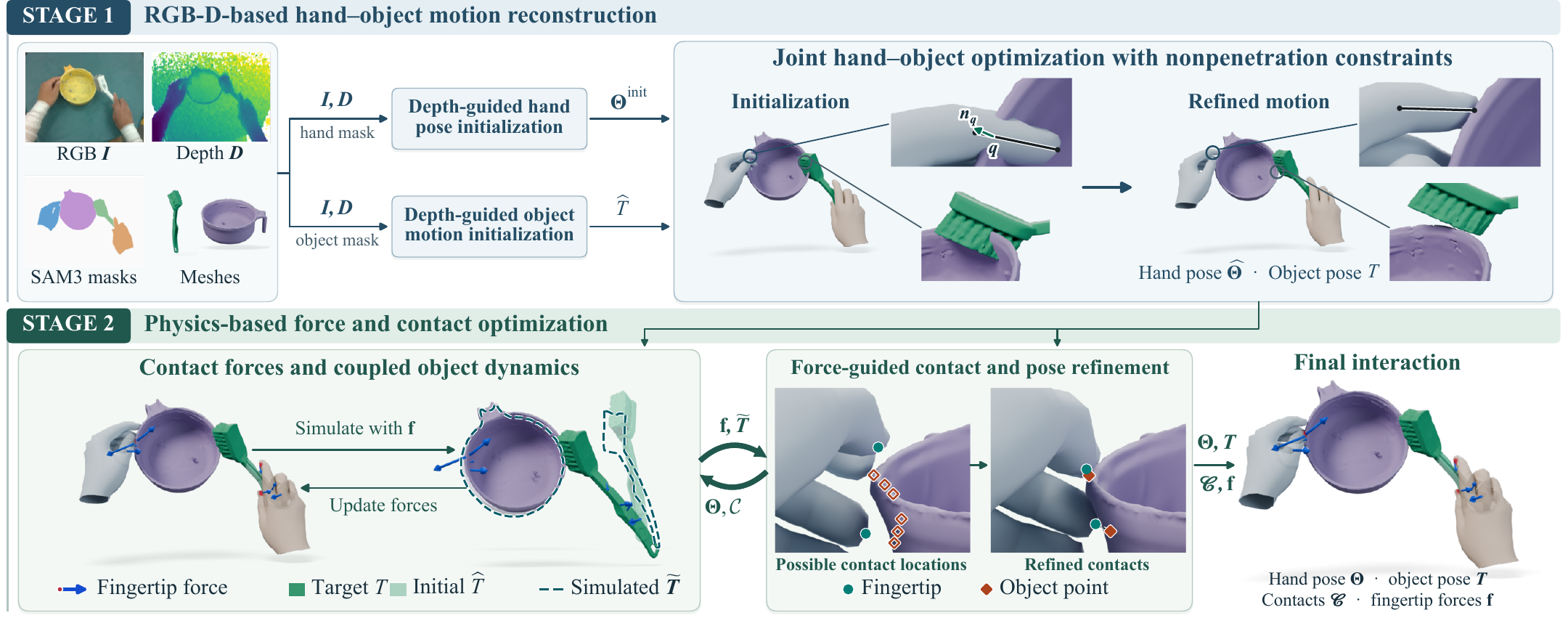}
  \caption{\textbf{Method overview.} Stage~1 jointly refines hand and object motion with nonpenetration constraints. The finger bone axis guides the correction at thin walls, while the object contact term penalizes overlap between the brush and bowl. Stage~2 fits fingertip forces by comparing simulated and reconstructed object motion. The remaining motion error guides adjustments to contact locations and hand poses, and the forces are then re-estimated at the updated contacts.}
  \label{fig:method_pipeline}
\end{figure*}

Figure~\ref{fig:method_pipeline} summarizes our framework. Given a single-view RGB-D video, its camera calibration, and the metric meshes of the manipulated objects, \methodname recovers the hand and object trajectories, fingertip contacts, and contact forces over the sequence. We first initialize the hand and object poses from RGB-D observations and jointly refine their motion over time (Sec.~\ref{sec:motion_reconstruction}). Using the reconstructed motion as a visual reference, the physical stage refines hand poses and contact locations and estimates fingertip forces (Sec.~\ref{sec:force_contact}).

\subsection{RGB-D-Based Hand--Object Motion Reconstruction}
\label{sec:motion_reconstruction}
At frame $t$, we observe RGB $I_t$, metric depth $D_t$, intrinsics $K_t$, and SAM3~\cite{carion2025sam3} instance masks. For each object $o$, we are given its metric mesh $\mathcal M_o$. We estimate its camera-space pose $T_{o,t}=(R_{o,t},p_{o,t})\in SE(3)$ from the RGB-D sequence. For each hand $h$, we use the left or right MANO model~\cite{romero2017mano} with fixed shape $\mathbf b_h\in\mathbb R^{10}$. The hand pose $\Theta_{h,t}$ comprises global orientation $R_{h,t}\in SO(3)$, camera-space wrist position $w_{h,t}\in\mathbb R^3$, and local rotations $\theta_{h,t}$ at 15 finger joints. For this shape, $J_j(\Theta_{h,t})$ gives the position of joint $j$. For moving cameras, we use the camera-to-world trajectory $C_t$ estimated by DROID-W~\cite{li2026droidw} to express the refined motion in world coordinates.

\subsubsection{Depth-Guided Object Motion Initialization}
\label{sec:object_tracking}
We use FoundationPose~\cite{wen2024foundationpose} to initialize each object trajectory from the RGB-D sequence. To follow rapid motion, we shift the previous pose by the displacement of the masked depth centroid before refining it against the current frame. If tracking is lost during occlusion, we re-register the object when it becomes visible and score the candidate poses by mask overlap and depth agreement.

\subsubsection{Depth-Guided Hand Pose Initialization}
\label{sec:hand_reconstruction}
We use WiLoR~\cite{potamias2025wilor} to initialize the wrist pose and finger rotations from each RGB hand crop. After transferring the rotations to the corresponding left or right MANO model, we correct wrist depth using the trimmed mean difference between observed and rendered depths within the hand mask. Along each track, we weight the Huber data term~\cite{huber1964robust} by valid depth coverage and penalize changes in depth velocity to bridge gaps in the measurements. We back-project the wrist pixel at the fitted depth to obtain $w_{h,t}$.

\subsubsection{Joint Hand--Object Optimization with Nonpenetration Constraints}
\label{sec:contact_cooptimization}
Let $X$ collect all hand poses $\Theta_{h,t}$ and object poses $T_{o,t}$ in a temporal window. We jointly refine $X$ by minimizing the weighted reconstruction loss $\mathcal L_{\mathrm{rec}}$:
\begin{multline}
\mathcal L_{\mathrm{rec}}(X)=
\lambda_{\mathrm{ICP}}\mathcal L_{\mathrm{ICP}}
+\lambda_{\mathrm{sil}}\mathcal L_{\mathrm{sil}}
+\lambda_{\mathrm{prior}}\mathcal L_{\mathrm{prior}}\\
{}+\lambda_{\mathrm{temp}}\mathcal L_{\mathrm{temp}}
+\lambda_{\mathrm{self}}\mathcal L_{\mathrm{self}}
+\lambda_{\mathrm{OO}}\mathcal L_{\mathrm{OO}}
+\lambda_{\mathrm{HO}}\mathcal L_{\mathrm{HO}}.
\end{multline}
Following InteractionFusion~\cite{zhang2019interactionfusion}, we align visible hand and object surfaces to the observed depth with projective ICP ($\mathcal L_{\mathrm{ICP}}$) and their projected contours to the instance masks ($\mathcal L_{\mathrm{sil}}$). The pose prior $\mathcal L_{\mathrm{prior}}$ keeps hand rotations and object poses close to their initial estimates, while $\mathcal L_{\mathrm{temp}}$ smooths hand and object motion over time. A self-collision penalty $\mathcal L_{\mathrm{self}}$ discourages overlap between capsules on different fingers~\cite{tagliasacchi2015articulated}. After visual alignment, we activate the object--object and hand--object nonpenetration terms, $\mathcal L_{\mathrm{OO}}$ and $\mathcal L_{\mathrm{HO}}$, and jointly refine all poses.

\paragraph{Object--object interaction.}
We sample $N_i$ points on the surface of object $i$ in proportion to triangle area and reuse them across frames. Let $\mathbf p_i^{(k)}\in\mathbb R^3$ denote the coordinates of the $k$th sample in object $i$'s local frame. At frame $t$, we transform each point into object $j$'s local frame and evaluate its signed distance $d_j$ to the surface of $j$ (positive outside). The directional loss is
\begin{equation}
\mathcal L_{i\to j}^{t}=\frac{1}{N_i}\sum_{k=1}^{N_i}
\left[\phi_\epsilon\!\left(d_j(T_{j,t}^{-1}T_{i,t}\mathbf p_i^{(k)})+\tau\right)\right]^2.
\end{equation}
The smoothed hinge $\phi_\epsilon$ activates near the tolerance $\tau$ and gives a quadratic loss for deeper penetrations (see supplement). The directional loss $\mathcal L_{i\to j}^{t}$ measures penetration of samples from $i$ into $j$. We average it with the reverse loss $\mathcal L_{j\to i}^{t}$ for each pair.
\begin{equation}
\mathcal L_{\rm OO}=\sum_t\sum_{i<j}\tfrac12\bigl(\mathcal L_{i\to j}^{t}+\mathcal L_{j\to i}^{t}\bigr).
\end{equation}

\paragraph{Hand--object interaction.}
We approximate finger segments by capsules of fitted radius $\rho$ and the palm by a convex polytope. As above, we sample both surfaces. The directional losses $\mathcal L_{h\to o}^{t}$ and $\mathcal L_{o\to h}^{t}$ average squared residuals $r_k^2$ over hand and object samples, respectively, weighted by the surface area each sample represents. Inactive samples have zero residual:
\begin{equation}
\mathcal L_{\rm HO}=\sum_t\sum_{h,o}\tfrac12\bigl(\mathcal L_{h\to o}^{t}+\mathcal L_{o\to h}^{t}\bigr).
\end{equation}

\begin{figure}[t]
\centering
\includegraphics[width=0.94\linewidth]{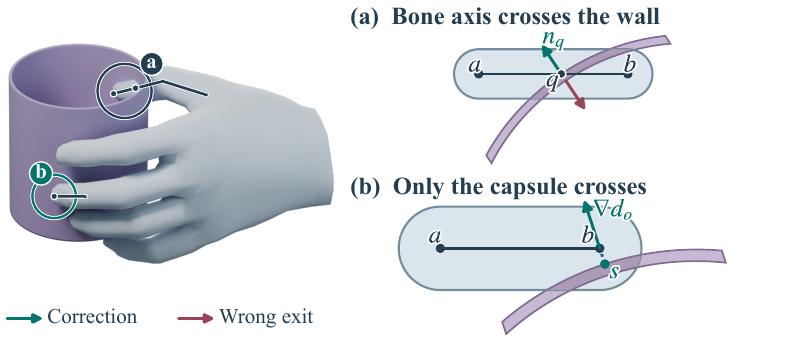}
\caption{\textbf{Resolving finger penetration at thin walls.} In both panels, the bone axis runs from parent joint $a$ to its child $b$. (a) The thumb axis crosses the wall at $q$. The entry-face normal guides the thumb back toward the side it entered from. (b) Only the middle-finger capsule crosses the wall. The object SDF gradient at $b$ identifies the exit direction.}
\label{fig:entryside_method}
\end{figure}

A finger penetrating a thin wall can be closer to the far face than to the face it crossed. Following the nearest outward normal would then push it farther through the wall. We select the contact face using the capsule axis from parent joint $a$ to child joint $b$. If the axis enters the wall at $q$ (Fig.~\ref{fig:entryside_method}a), the outward normal $n_q$ of the first entry face defines the residual:
\begin{equation}
 r_k=\phi_\epsilon\!\left(n_q^\top(b-q)-\rho+\tau\right).
\end{equation}
Here $n_q^\top(b-q)$ measures the child joint's signed distance to the entry plane. Subtracting the radius measures how far the capsule extends past that plane, so the correction accounts for the thickness of the finger.
When only the capsule overlaps the wall, the axis has no entry point (Fig.~\ref{fig:entryside_method}b). The object SDF gradient at the axis indicates the side occupied by the finger, allowing us to reject a nearby face whose outward normal points the other way. If the face is ambiguous, we reuse a compatible plane stored in object coordinates. Without one, we omit the residual. Switching sides requires consistent evidence over several frames. The palm uses the bidirectional signed-distance penalty above.

An occluded finger can pull a well-tracked object away from its depth observations. We therefore attenuate the object's contact-induced update as its relative visual confidence $\widetilde c_{ho,t}\in[0,1]$ increases. This confidence is the temporally smoothed object share of squared ICP support for the pair, with each entity's support measured in accepted correspondences per degree of freedom.

\subsection{Physics-Based Force and Contact Optimization}
\label{sec:force_contact}
We augment the reconstructed states $X=\{\Theta,T\}$ of
Sec.~\ref{sec:motion_reconstruction} with contact correspondences $\mathcal C$
and nonnegative impulse coefficients $\alpha$. Here $\widehat T$ denotes
the tracked object initialization, $T$ the refined object trajectory, and
$\widetilde T$ its simulated counterpart; $\widehat J$ provides the visual
hand-joint references. The stages use the following loss blocks:
\begin{equation}
\begin{aligned}
 \mathcal L
 ={}&\lambda_{\mathrm{dyn}}\mathcal L_{\mathrm{dyn}}
    +\lambda_{\mathrm{contact}}\mathcal L_{\mathrm{contact}}\\
   &+\lambda_{\mathrm{obs}}\mathcal L_{\mathrm{obs}}
    +\lambda_{\mathrm{prior}}\mathcal L_{\mathrm{prior}}
    +\lambda_{\mathrm{temp}}\mathcal L_{\mathrm{temp}}.
\end{aligned}
\label{eq:joint_physical_objective}
\end{equation}
$\mathcal L_{\mathrm{prior}}$ combines pose, self-collision, and anatomical
terms. $\mathcal L_{\mathrm{temp}}$ regularizes hand motion, force changes,
and object-local contact points and normals, allowing sliding.
Coefficients are given in the supplement.

\subsubsection{Contact Forces and Coupled Object Dynamics}
\label{sec:contact_dynamics}
Each contact has a finger--object assignment and points on both surfaces.
Updates retain the assignment and hand-mesh correspondence while adjusting
the object-surface point. MANO~\cite{romero2017mano} determines the hand
point; fingertip forces act externally on the object.

Rays around the inward normal approximate the friction cone.
At contact $i$ on object $o(i)$, the world-space force is
$\widetilde{\mathbf f}_{t,i}(s)=\widetilde R_{o(i)}(s)
D_i(\mathcal C_t,s)\alpha_{t,i}/\Delta t_t$.
The simulated rotation $\widetilde R_{o(i)}(s)$ maps the object-local cone
rays in $D_i$ to world coordinates at time $s$. The coefficients
$\alpha_{t,i}$ are shared across the frame interval $\Delta t_t$.
This represents normal and tangential forces with normal-load limits per finger and hand.

All manipulated objects evolve in one MuJoCo~\cite{todorov2012mujoco} scene:
\begin{equation}
\begin{aligned}
 M(q)\dot\nu+h(q,\nu)
 ={}&G_h(q,\mathcal C)\mathbf f\\
   &+J_{\mathrm{oo}}^{\top}\lambda_{\mathrm{oo}}+J_e^{\top}\lambda_e.
\end{aligned}
\label{eq:coupled_dynamics}
\end{equation}
Here $q,\nu$ are object configurations and velocities; $M,h$ describe
inertia and bias forces including gravity. $G_h$ maps fingertip forces
to generalized forces. MuJoCo solves for object--object and support
reactions $\lambda_{\mathrm{oo}},\lambda_e$, mapped through their contact
Jacobians; paired reactions act through the respective moment arms.

Integrating Eq.~\eqref{eq:coupled_dynamics} under the fingertip forces yields
simulated object states $\widetilde x_{o,t}$. The reconstructed state
$x_{o,t}(T)$ comprises position, orientation, and linear and angular
velocities derived from $T$. The dynamics loss
$\mathcal L_{\mathrm{dyn}}=\tfrac12\sum_{t,o}\omega_t
\|r_{\mathrm{dyn}}(\widetilde x_{o,t},x_{o,t}(T))\|^2$
compares these states using separate residual scales. Motion transmitted
through contact thus helps constrain the fingertip forces.

\subsubsection{Force-Guided Contact and Pose Refinement}
\label{sec:contact_refinement}
Contact locations determine moment arms and friction directions.
The reconstructed pose $T_{o(i),t}$ maps object-local points into world
coordinates. For surface point $c_{t,i}$ and outward normal $n_{t,i}$,
the contact-cost target is $\bar c_{t,i}=c_{t,i}-\epsilon_g n_{t,i}$.
Its gap to the hand point is
$g_{t,i}=v_i(\Theta_t)-T_{o(i),t}\bar c_{t,i}$; forces act at $c_{t,i}$.
The contact loss $\mathcal L_{\mathrm{contact}}$ penalizes separation more
strongly under higher loads, following Physical Interaction~\cite{hu2022physicalinteraction}.
\begin{equation}
\begin{aligned}
 \mathcal L_{\mathrm{contact}}
 ={}&\frac{w_n}{2}\sum_{t,i,s}\pi_{t,s}
 \left(\frac{a_{t,i}}{a_0}\right)^2
 \left\|\frac{g_{t,i}(s)}{d_0}\right\|^2\\
 &+\lambda_{\mathrm{acq}}\sum_{t,i,s}\frac{\omega_t\pi_{t,s}}{N_t}
 \psi\!\left(\frac{\|g_{t,i}(s)\|}{d_0}\right).
\end{aligned}
\label{eq:load_contact}
\end{equation}
Here $a_{t,i}=\mathbf1^\top\alpha_{t,i}/\Delta t_t$ is the load proxy, and
$a_0,d_0$ normalize load and distance. Here $\pi_{t,s}$ are substep
quadrature weights, $\omega_t$ equal interval weights, and $N_t$ the
contact count. The supplement defines $w_n$, $\epsilon_g$, and $\psi$.

Contacts on a hand share MANO parameters; those on an object share its
rigid pose. Contact and hand-pose updates evaluate gaps using $T$.
During force estimation, the load--gap and attraction penalties
instead use simulated object states, with hand geometry fixed. Both
evaluations use the same current contact correspondences.

The observation loss $\mathcal L_{\mathrm{obs}}$ anchors hand joints and
object poses to their visual references $\widehat J$ and $\widehat T$.
\begin{equation}
\begin{aligned}
 \mathcal L_{\mathrm{obs}}
 ={}&\frac12\sum_{t,h,j}w^h_{h,j,t}
 \|J_j(\Theta_{h,t})-\widehat J_{h,j,t}\|^2_{\Sigma_{h,j}^{-1}}\\
 &+\sum_{t,o}w^o_{o,t}d_o^2(T_{o,t},\widehat T_{o,t}).
\end{aligned}
\label{eq:visual_anchoring}
\end{equation}
Here $J_j$ gives reconstructed joints, $\Sigma_{h,j}$ sets their error
scales, and $d_o$ combines normalized translation and rotation errors.
Hand weights $w^h$ follow joint visibility. The object term is the tracking
prior of Sec.~\ref{sec:contact_cooptimization}, with $w^o=\gamma_{o,t}$;
its metric and confidence-weighted pose update are specified in the supplement.

\subsubsection{Multi-Frame Alternating Optimization}
\label{sec:physical_objective}
Adjacent intervals share frame states. We linearize motion residuals
with respect to cone impulses and evaluate constrained updates in MuJoCo.
Reference motion provides force and torque seeds for contact
selection and hand-pose fitting, followed by force re-estimation.
The supplement specifies the subproblems, weights, geometric constraints,
and acceptance criteria.

\subsubsection{Force-Driven Replay and Cross-Engine Transfer}
\label{sec:force_replay_transfer}
Recovered contacts and forces reproduce the MuJoCo trajectory by continuous
replay from the initial state. For PyBullet and Isaac Sim, we adapt forces
while retaining physical parameters and application-point sequences.
Independent replay uses no intermediate state resets. The supplement and
Sec.~\ref{sec:force_evaluation} provide adaptation details and evaluation.

\section{Experiments}
\label{sec:experiments}
% CompactExperimentBox is defined in preamble.tex.
% Verify that the preview compiler resolves distinct regular and bold fonts.
\begingroup
\normalfont\fontsize{8}{9.2}\selectfont
\edef\ExperimentRegularFont{\fontname\font}
\bfseries\selectfont
\edef\ExperimentBoldFont{\fontname\font}
\ifx\ExperimentRegularFont\ExperimentBoldFont
\PackageError{experiments}{Bold and regular resolve to the same font}{Check font availability.}
\fi
\endgroup

We first evaluate the reconstructed hands, object trajectories, and the corresponding grasps using the RGBD HOI datasets DexYCB~\cite{chao2021dexycb} and TACO~\cite{liu2024taco} (Sec.~\ref{sec:reconstruction_results}). Then we compare fingertip forces with sensor measurements and test whether the recovered forces reproduce object motion in simulation (Sec.~\ref{sec:force_evaluation}). Finally, we isolate the effects of the key components of our technique, including tracking, joint refinement, and contact modeling (Sec.~\ref{sec:ablations}). The supplementary material details the reconstruction and physics solvers.

\subsection{Hand--Object Reconstruction}
\label{sec:reconstruction_results}
\label{sec:experimental_setup}
\paragraph{Evaluation protocol.}
We evaluate 177 DexYCB sequences spanning all 10 subjects and 20 object categories, and 228 first-person TACO sequences of bimanual, two-object manipulation. We report ADD/ADD-S AUC and translation and rotation errors for objects. For hands, we report MPJPE and its per-frame (PA) and window-aligned (WA) variants. Acceleration error measures motion fidelity, while penetration depth (PD) and intersection volume (IV) measure hand--object and object--object overlap.

For object tracking, we compare with FoundationPose (FP)~\cite{wen2024foundationpose}, ICG~\cite{stoiber2022icg}, and CAD-free Point2Pose~\cite{lin2026point2pose}. For hand reconstruction, we compare with RGB-based HaWoR and Dyn-HaMR~\cite{zhang2025hawor,yu2025dynhamr}. For hand--object reconstruction, we compare with the single-object methods HOTrack~\cite{chen2023hotrack} and AGILE~\cite{shi2026agile}. We evaluate both only on DexYCB because TACO requires joint tracking of two objects. FP, ICG, HOTrack, and AGILE receive metric object meshes and measured depth.

For each dataset, all object trackers follow the same initialization protocol. On DexYCB, we use the annotated object pose at the start of tracking. HOTrack also requires a hand pose, so we start scoring it after successful initialization. On TACO, we refine the first-frame object annotation against the observed mask and depth and share the resulting pose across trackers. For evaluation, we score all methods directly in camera coordinates on DexYCB. On TACO, we align each completed trajectory to the egocentric annotations with a single rigid transform to account for offsets in the projected annotations. We evaluate the full sets at 30\,Hz. For the AGILE comparison, we score both methods on the same valid frames at 6\,Hz across 100 DexYCB sequences covering all subjects and object categories. The supplement gives the alignment and metric definitions.

\begin{figure}[t]
  \centering
  \includegraphics[width=\linewidth]{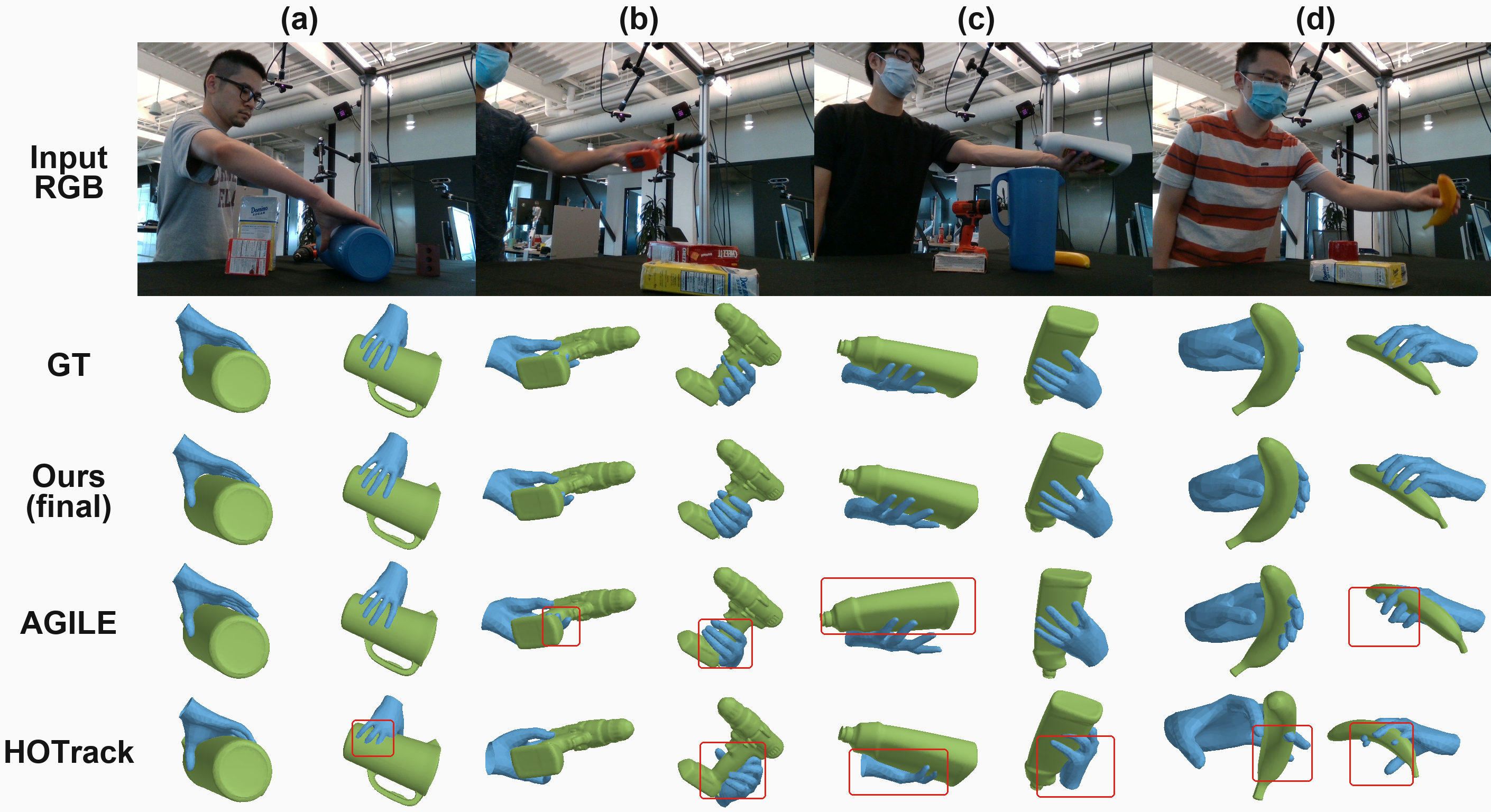}
  \caption{Hand--object reconstruction on DexYCB. Each column shows one example, with rows showing the input RGB, ground truth, our final result, AGILE, and HOTrack. Each reconstruction is rendered from two viewpoints shared by all methods. Red boxes highlight finger penetration, object-pose discrepancies, and mismatched grasp configurations in the baselines.}
  \label{fig:dexycb_grasps}
\end{figure}

\begin{table}[t]
\centering
\caption{Object tracking and hand reconstruction on TACO.}
\label{tab:taco_results}
{\fontsize{8}{9.2}\selectfont\setlength{\tabcolsep}{1pt}
\begin{tabular*}{\columnwidth}{@{\extracolsep{\fill}}lrrrrrrr@{}}
\toprule
\textbf{Objects} & ADD$\uparrow$ & ADD-S$\uparrow$ & $T\downarrow$ & $R\downarrow$ & $A\downarrow$ & PD$_{\rm OO}\downarrow$ & IV$_{\rm OO}\downarrow$\\
\midrule
FP & 81.65 & 92.54 & 11.46 & 15.77 & 3.09 & 0.33 & 0.14\\
ICG & 77.43 & 89.04 & 15.84 & 20.13 & 3.60 & 0.85 & 0.93\\
Point2Pose & \underline{84.57} & \underline{92.62} & \underline{11.30} & \underline{12.28} & 5.98 & 0.38 & 0.25\\
\midrule
\textbf{Ours} & \textbf{86.31} & \textbf{95.52} & \textbf{7.37} & \textbf{11.75} & \textbf{1.20} & \underline{0.12} & \underline{0.02}\\
Ours (replay) & 76.52 & 90.02 & 14.28 & 17.97 & \underline{1.83} & \textbf{0.03} & \textbf{0.00}\\
\bottomrule
\end{tabular*}\par
\smallskip
\begin{tabular*}{\columnwidth}{@{\extracolsep{\fill}}lrrrrrr@{}}
\toprule
\textbf{Hand pose} & MPJPE$\downarrow$ & PA$\downarrow$ & WA$\downarrow$ & $A\downarrow$ & PD$\downarrow$ & IV$\downarrow$\\
\midrule
HaWoR & \underline{44.00} & \underline{8.62} & \underline{27.31} & 14.07 & -- & --\\
Dyn-HaMR & 57.09 & 9.58 & 29.85 & \underline{9.69} & -- & --\\
\midrule
\textbf{Ours} & \textbf{22.18} & \textbf{8.22} & \textbf{15.43} & \textbf{5.07} & 2.54 & 0.33\\
\bottomrule
\end{tabular*}\par
}
{\scriptsize\raggedright ADD/ADD-S report AUC (\%). $T$ and $R$ denote translation (mm) and rotation ($^\circ$) errors. $A$ is acceleration error (m/s$^2$). MPJPE, PA, WA and PD are in mm, and IV in cm$^3$. PD$_{\rm OO}$ and IV$_{\rm OO}$ measure overlap between the two objects. Best and second best within each panel are bold and underlined.\par}
\end{table}

\begin{table}[t]
\centering
\caption{Object tracking and hand reconstruction on DexYCB.}
\label{tab:dexycb_results}
{\fontsize{8}{9.2}\selectfont\setlength{\tabcolsep}{1pt}
\begin{tabular*}{\columnwidth}{@{\extracolsep{\fill}}lrrrrr@{}}
\toprule
\textbf{Object tracking} & ADD$\uparrow$ & ADD-S$\uparrow$ & $T\downarrow$ & $R\downarrow$ & $A\downarrow$\\
\midrule
FP & 73.47 & 79.28 & 62.37 & 21.30 & 3.79\\
ICG & 72.57 & 83.16 & 41.84 & 25.89 & 3.62\\
Point2Pose & 74.07 & 80.10 & 55.12 & 22.82 & 7.30\\
HOTrack$^*$ & 79.83 & 92.76 & \underline{13.31} & 22.19 & 4.80\\
AGILE$^*$ & 54.59 & 69.20 & 81.43 & 26.55 & --\\
\midrule
\textbf{Ours} & \textbf{84.87} & \textbf{93.25} & \textbf{12.50} & \underline{15.73} & \textbf{1.41}\\
Ours (replay) & 73.38 & 81.52 & 24.58 & 18.01 & \underline{2.22}\\
S1 w/o joint opt. & \underline{84.64} & \underline{92.85} & 13.37 & \textbf{15.25} & 3.57\\
Ours (AGILE)$^*$ & 85.73 & 94.62 & 10.02 & 14.82 & --\\
\bottomrule
\end{tabular*}\par
\smallskip
\begin{tabular*}{\columnwidth}{@{\extracolsep{\fill}}lrrrrrr@{}}
\toprule
\textbf{Hand pose} & MPJPE$\downarrow$ & PA$\downarrow$ & WA$\downarrow$ & $A\downarrow$ & PD$\downarrow$ & IV$\downarrow$\\
\midrule
HaWoR & 38.17 & 5.51 & 24.53 & 17.80 & -- & --\\
Dyn-HaMR & 104.28 & 6.14 & 23.43 & 8.78 & -- & --\\
HOTrack$^*$ & 28.85 & 13.03 & 24.89 & 10.36 & 9.40 & 9.75\\
AGILE$^*$ & 20.39 & 4.46 & 16.75 & -- & 6.79 & 8.00\\
\midrule
\textbf{Ours} & \underline{20.88} & \textbf{4.57} & \underline{15.71} & 5.01 & \underline{1.94} & \underline{1.11}\\
S1 (no physics) & 21.63 & \underline{4.58} & 16.10 & \underline{4.72} & \textbf{1.29} & \textbf{0.92}\\
S1 w/o nonpen. & \textbf{18.12} & \textbf{4.57} & \textbf{14.29} & \textbf{4.16} & 6.82 & 6.80\\
S1 w/o joint opt. & 27.52 & 4.73 & 22.85 & 8.07 & 6.56 & 6.91\\
Ours (AGILE)$^*$ & 17.06 & 4.35 & 12.59 & -- & 2.17 & 1.01\\
\bottomrule
\end{tabular*}\par
}
{\scriptsize\raggedright $^*$HOTrack averages cover successful sequences only (141/177 for objects, 79.66\%, and 115/177 for hands, 64.97\%). AGILE and Ours (AGILE) share the same 100 sequences at 6\,Hz and are excluded from ranking. S1 denotes Stage~1. Units as in Table~\ref{tab:taco_results}.\par}
\end{table}

\paragraph{Object tracking.}
\methodname recovers object poses more accurately and with lower acceleration error than the object trackers FP, ICG, and Point2Pose on both datasets (Tables~\ref{tab:taco_results} and~\ref{tab:dexycb_results}).
On TACO, \methodname also reduces inter-object penetration relative to all three trackers (see supplemental ablations).
\methodname outperforms HOTrack on all five object tracking metrics, with our results averaged over the full DexYCB set and HOTrack's over its successful tracks. Compared with AGILE on the same frames, \methodname estimates both object position and orientation more accurately.

\paragraph{Hand pose and interaction.}
\methodname outperforms HaWoR and Dyn-HaMR on all four hand motion metrics across both datasets (Tables~\ref{tab:taco_results} and~\ref{tab:dexycb_results}). Its full-set results also surpass HOTrack's averages over successful tracks on all six hand and interaction metrics. On the same frames as AGILE, \methodname improves all five reported metrics, with larger gains in hand placement and motion than in per-frame aligned pose, and less hand--object penetration (Fig.~\ref{fig:dexycb_grasps}).

\begin{figure}[t]
  \centering
  \includegraphics[width=\linewidth]{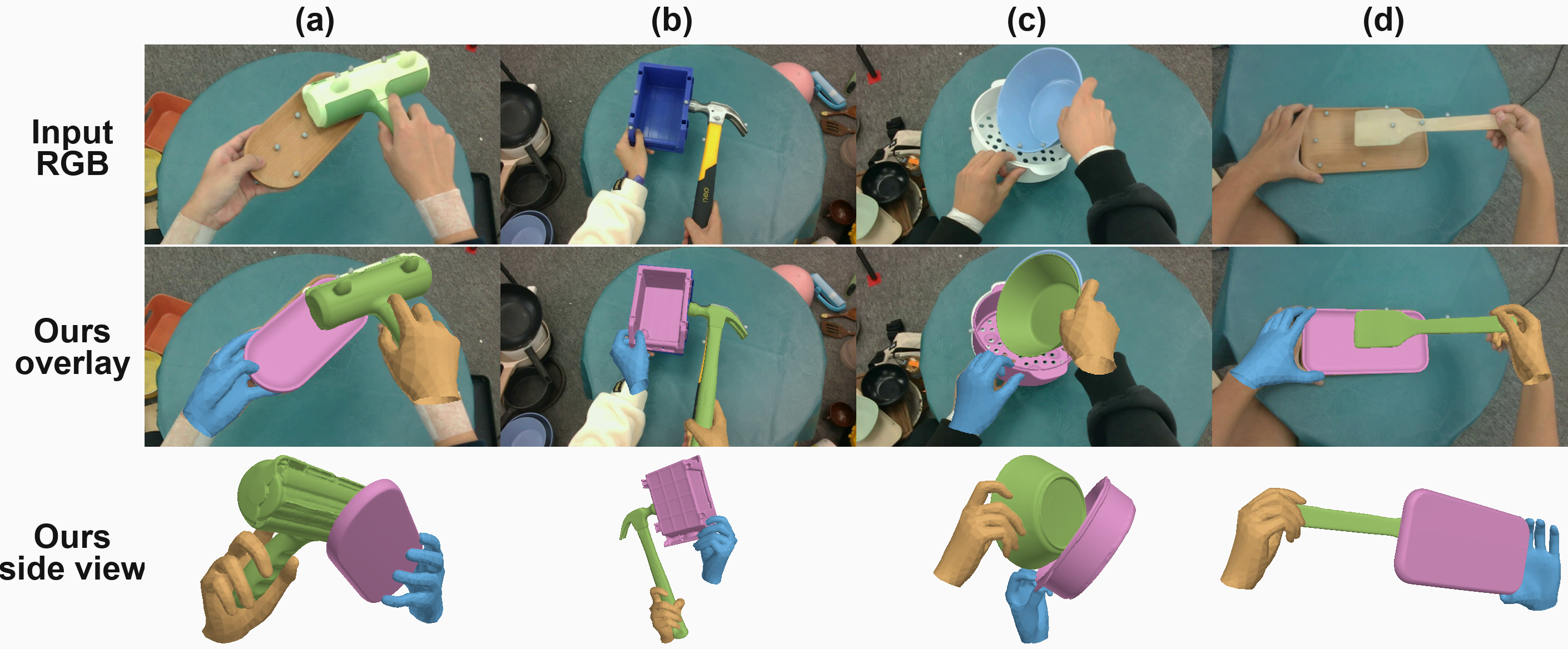}
  \caption{Bimanual two-object reconstruction on TACO. Each column shows one interaction from the input view (RGB and mesh overlay) and an alternate view.}
  \label{fig:taco_contacts}
\end{figure}

\subsection{Hand--Object Force Reconstruction and Replay}
\label{sec:force_evaluation}

\begin{table}[t]
\centering
\caption{\textbf{Measured hand--object force accuracy.} Sequence-balanced normal-force errors on 32 sequences. Best in bold; second best underlined.}
\label{tab:force_accuracy}
{\small\setlength{\tabcolsep}{4pt}\renewcommand{\arraystretch}{1.08}
\begin{tabular*}{\columnwidth}{@{\extracolsep{\fill}}lrr@{}}
\toprule
Method & MAE (N) $\downarrow$ & RMSE (N) $\downarrow$ \\
\midrule
EgoTac~\cite{zhang2026egotac} & 1.473 & 2.236 \\
EgoPHI~\cite{ilic2026egophi} & 1.396 & 2.132 \\
HOPE~\cite{jeon2026hope} & \underline{1.326} & \underline{2.063} \\
\midrule
Ours & \textbf{1.002} & \textbf{1.523} \\
\bottomrule
\end{tabular*}}
\end{table}

\begin{figure*}[t]
\centering
% Stack the two panels separately so the video strip can be smaller than the curves.
\vbox{\offinterlineskip
  \hbox to \textwidth{\hfil\includegraphics[width=0.66\textwidth,trim=0bp 105bp 0bp 0bp,clip]{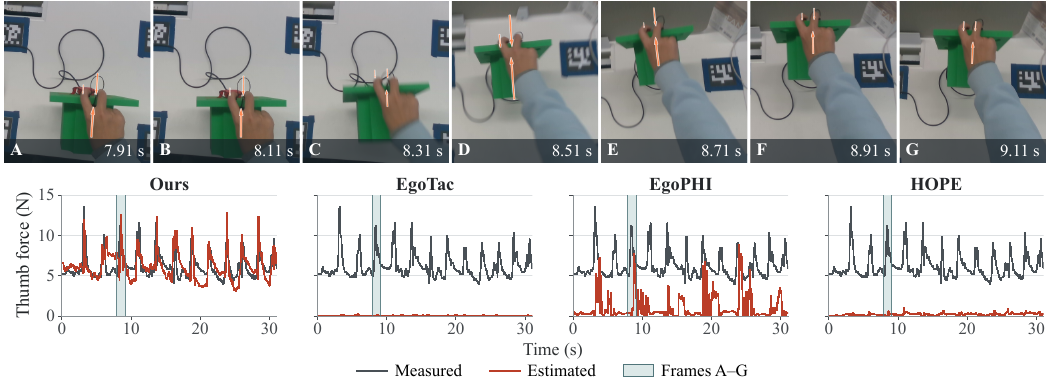}\hfil}
  \kern 1pt
  \hbox to \textwidth{\hfil\includegraphics[width=0.82\textwidth,trim=0bp 0bp 0bp 84bp,clip]{figures/force_curves.pdf}\hfil}}
\caption{\textbf{Force reconstruction under changing loads.} Arrows show estimated normal forces in frames A--G (shaded). Curves compare measured and estimated thumb forces over 931 intervals (31\,s). See the supplement for all three fingers.}
\label{fig:force_curves}
\end{figure*}

\paragraph{Measured-force accuracy.}
We recorded 32 sequences spanning six objects and seven interaction settings, measuring fingertip forces with PaXini PX6AX GEN4 tactile sensors~\cite{paxiniGen4}. \methodname estimates normal and tangential fingertip forces using object masses and friction coefficients, with sensor readings reserved for evaluation. The three open-source baselines estimate only normal loads, so we compare normal forces on the same valid sensor intervals. Relative to HOPE, the strongest baseline, \methodname reduces MAE by 24.5\% and RMSE by 26.1\% (Table~\ref{tab:force_accuracy}). In a three-finger grasp, our estimates track repeated peaks in thumb force (Fig.~\ref{fig:force_curves}).

Across all 32 sequences, \methodname achieves a mean 3D force-vector error of 1.222\,N, versus 1.486\,N with its tangential components set to zero (see supplement). We also compare with a GPT-6 Astra~\cite{openai2026gpt6astra} agent on eight sequences. It reconstructs interactions in MuJoCo from RGB-D images, masks, and meshes and reads simulated contact forces, using the same masses, inertias, and friction coefficients as \methodname. \methodname has lower Euclidean error on all eight sequences, averaging 1.051\,N versus 3.874\,N.

\paragraph{Force-driven replay.}
To test whether the recovered fingertip forces reproduce the observed object motion, we apply them continuously from the initial state in MuJoCo. Tables~\ref{tab:taco_results} and~\ref{tab:dexycb_results} report translation ($T$) and rotation ($R$) errors against the dataset annotations for both replay (\emph{Ours (replay)}) and RGB-D tracking (\emph{Ours}). The supplement compares replay with the visual trajectories used to fit the forces.

The tactile measurements test the load on individual fingers. Replay tests their combined effect on the object, since an error in where a force acts can change the rotation even when its magnitude agrees with the sensor.
\begin{table}[t]
\centering
\caption{\textbf{Cross-engine transfer.} Median translation RMSE against MuJoCo over 21 sequences, combining both objects per sequence, and valid insertion counts.}
\label{tab:engine_transfer}
{\small\setlength{\tabcolsep}{2.5pt}\renewcommand{\arraystretch}{1.05}
\begin{tabular*}{\columnwidth}{@{\extracolsep{\fill}}llrr@{}}
\toprule
Engine & Forces & RMSE (mm)$\downarrow$ & Insertion$\uparrow$ \\
\midrule
MuJoCo & Source & -- & 20/21 \\
\midrule
PyBullet & Direct & 3563.90 & 0/21 \\
 & Adapted & \textbf{5.12} & \textbf{19/21} \\
\midrule
Isaac Sim & Direct & 1432.69 & 0/21 \\
 & Adapted & \textbf{13.37} & \textbf{19/21} \\
\bottomrule
\end{tabular*}}
\end{table}

We next test transfer to PyBullet and Isaac Sim on the rod--hole insertion sequences from the measured-force set, also used in the contact ablation below. Directly applying the MuJoCo forces fails to reproduce insertion in either engine. Refitting the forces to the MuJoCo trajectories with contact locations and physical parameters fixed restores insertion in most sequences in both engines (Table~\ref{tab:engine_transfer}). Normal-force MAE after refitting remains close to MuJoCo's 1.020\,N (1.007\,N in PyBullet and 1.022\,N in Isaac Sim). The supplement shows the final frames of all PyBullet replays.

\subsection{Ablation Studies}
\label{sec:ablations}
\begin{figure*}[t]
  \centering
  \includegraphics[width=0.82\textwidth]{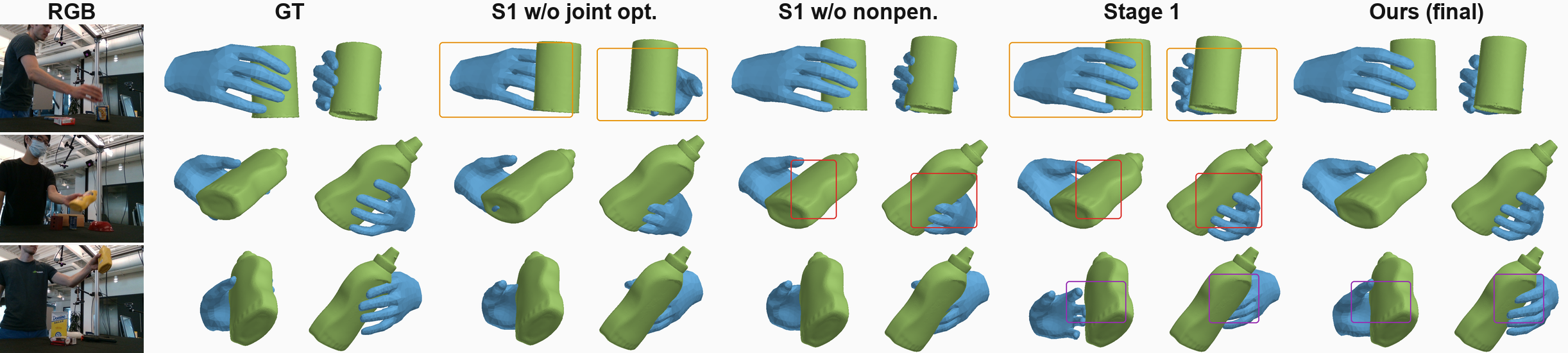}
  \caption{\textbf{DexYCB grasp ablations.} Paired views highlight changes in hand placement (orange), penetration (red), and fingertip contact (purple).}
  \label{fig:dexycb_ablations}
\end{figure*}
\paragraph{Object tracking initialization.}
Our enhanced FoundationPose tracker adds depth-based displacement updates and re-registration. Before joint optimization, it already yields lower object translation error than the original FoundationPose (S1 w/o joint opt.\ vs.\ FP in Table~\ref{tab:dexycb_results}).

\paragraph{Joint hand--object optimization.}
Refining the initial trajectories improves most object pose metrics, substantially reduces object acceleration error, and lowers camera-space hand joint error. The orange boxes in Fig.~\ref{fig:dexycb_ablations} show the resulting improvement in hand placement relative to the object. Within this optimization, we isolate the hand--object nonpenetration term by comparing S1 (no physics) with S1 w/o nonpen. Enabling the term reduces both penetration depth and volume, as illustrated by the finger in the red box. Correcting an intersecting finger can also move its joints farther from their annotations, explaining the increase in MPJPE.

\paragraph{Physical contact refinement.}
Physical refinement reduces both MPJPE and WA relative to Stage~1, with a small increase in hand--object overlap (Table~\ref{tab:dexycb_results}). Figure~\ref{fig:dexycb_ablations} (purple) shows the fingertip regaining contact.

\paragraph{Object--object contact during force fitting.}
We refit fingertip forces on the 21 rod--hole insertions with rod--base contact response enabled or disabled, keeping contact locations and solver settings fixed. Both force sets are then replayed in MuJoCo with rod--base contact response enabled. Although normal-force MAE is nearly unchanged (1.020 versus 1.021\,N), valid final insertions fall from 20/21 to 0/21, showing that matching fingertip loads alone does not reproduce the interaction between objects.

\section{Conclusion}
\label{sec:conclusion}
We presented \methodname for reconstructing hand--object motion and fingertip forces from single-view RGB-D video. Temporal joint optimization combines visual alignment with robust nonpenetration constraints, using hand skeletal geometry to resolve correction directions at thin surfaces. Coupled inverse dynamics accounts for loads transmitted between manipulated objects and guides joint refinement of contacts, poses, and forces. In the experiments, \methodname recovers more accurate hand and object trajectories while reducing both hand--object and inter-object penetration. Its fingertip force estimates match tactile measurements more closely than those of the baselines. These forces also reproduce the observed object motion in MuJoCo and can be adapted for replay in other physics engines.

\paragraph{Limitations.}
First, we require a metric-scale mesh and mass for each object, along with contact friction coefficients. RGB-D reconstruction methods such as BundleSDF~\cite{wen2023bundlesdf} can supply the mesh.

Second, \methodname currently supports only rigid objects. To track how an object's surface deforms under contact, future work could incorporate a local deformation field as in Physical Interaction~\cite{hu2022physicalinteraction}.

Third, we model each contacting fingertip with a single point force. This approximation omits forces exerted by the palm and pressure variations across the finger pad.

Finally, object motion alone cannot uniquely determine the forces at individual fingertips. Two opposing fingertips can squeeze harder while producing the same net force and torque. Our estimates therefore represent one physically feasible distribution and may differ from the forces actually applied. Nevertheless, their ability to reproduce object motion in simulation suggests they could serve as force targets for robot imitation learning.

\section*{LLM Usage Statement}
OpenAI Codex assisted us in drafting and editing the manuscript, checking citations, and clarifying our claims and conclusions. It also assisted with implementation, data preparation, experimental evaluation and analysis, and figure preparation. We provided the manuscript and supplementary material to OpenAI Codex to obtain reviewer-style feedback. An OpenAI GPT-6 Astra agent was also used as a simulation-based comparison, as described in Sec.~\ref{sec:force_evaluation}. Piper was used to synthesize narration for the supplementary video. We reviewed and edited all submitted content and take full responsibility for it.

{
    \small
    \bibliographystyle{ieeenat_fullname}
    \bibliography{./main}
    
}

% Optional: include only when preparing a combined main-and-supplementary PDF.
\clearpage
\appendix
\setcounter{equation}{0}
\setcounter{table}{0}
\setcounter{figure}{0}
\renewcommand{\theequation}{S\arabic{equation}}
\renewcommand{\thetable}{S\arabic{table}}
\renewcommand{\thefigure}{S\arabic{figure}}
\renewcommand{\theHequation}{supp.\arabic{equation}}
\renewcommand{\theHtable}{supp.\arabic{table}}
\renewcommand{\theHfigure}{supp.\arabic{figure}}
\maketitlesupplementary

\section{Hand--Object Motion Reconstruction Details}
\label{sec:supp_motion}

\subsection{Wrist depth fitting}

Within the hand mask at frame $t$, we retain pixels with valid rendered depth and measured depth between 200 and 2000\,mm. If at least 64 pixels qualify, we add the mean measured-minus-rendered depth difference between the 5th and 95th percentiles to the initial wrist depth. This gives the wrist depth observation $y_t$. Its weight $\omega_t$ is the number of valid pixel pairs divided by the visible rendered hand area in pixels.

Let $z_t$ be the fitted wrist depth in millimeters, $\mathcal V$ the frames with valid observations, and $\mathcal I$ the interior frames of a continuous track. With $\Delta_t$ the time from frame $t$ to $t+1$ and $m$ the median frame interval, the scaled change in depth velocity is
\begin{equation}
 a^z_t=m\left(\frac{z_{t+1}-z_t}{\Delta_t}
              -\frac{z_t-z_{t-1}}{\Delta_{t-1}}\right).
\end{equation}
Using a Huber loss $H_\delta$ with threshold $\delta=25$\,mm and smoothing weight $\lambda_z=1$, we solve
\begin{equation}
 \min_{\{z_t\}}\sum_{t\in\mathcal V}\omega_t H_\delta(z_t-y_t)
       +\frac{\lambda_z}{2}\sum_{t\in\mathcal I}(a^z_t)^2 .
\end{equation}
For a residual $e$, $H_\delta(e)=e^2/2$ when $|e|\leq\delta$ and $H_\delta(e)=\delta(|e|-\delta/2)$ otherwise. We initialize the fit by linear interpolation between observations and use iteratively reweighted least squares until the largest depth update falls below $0.001$\,mm or 50 iterations elapse. Beyond the first and last observations we extend the nearest fitted depth. With a single observation we use that depth throughout the track.

\subsection{Joint hand--object optimization}

\subsubsection{Residual terms}

\paragraph{Depth alignment ($\mathcal L_{\mathrm{ICP}}$).} For a visible model point $p$, we search its projected image neighborhood for a depth match $q$ and reject matches with incompatible normals or a 3D distance above 25\,mm. The hand residual is $\|q-p\|_2$. The rigid object residual is $n_p^\top(q-p)$, where $n_p$ is the unit model normal at $p$.

\paragraph{Silhouette alignment ($\mathcal L_{\mathrm{sil}}$).} A projected contour sample outside the mask is matched to the nearest foreground pixel. Conversely, an uncovered mask boundary sample is matched to the nearest visible model contour. For hands, the squared pixel distances in these two directions receive weights of $1.5$ and $1.5\times0.3125$, respectively.

\paragraph{Pose priors ($\mathcal L_{\mathrm{prior}}$).} Let $\ell(R,S)=\log(RS^\top)$ denote the rotation vector in radians from reference orientation $S$ to current orientation $R$. For hands, we use $\|\ell(R,S)\|_2^2$ with the initial WiLoR orientation as $S$ and separate weights for the wrist and fingers. For object $o$ at frame $t$, let $(R_{o,t},p_{o,t})$ denote its current pose, $(R^0_{o,t},p^0_{o,t})$ its tracked pose, and $\gamma_{o,t}$ its tracking confidence. With translations in millimeters, the tracking prior contributes
\begin{equation}
 \gamma_{o,t}\bigl(9\|p_{o,t}-p^0_{o,t}\|_2^2+
 22500\|\ell(R_{o,t},R^0_{o,t})\|_2^2\bigr).
\end{equation}

\paragraph{Temporal regularization ($\mathcal L_{\mathrm{temp}}$).} We use first- and second-order motion regularization as in articulated tracking~\cite{tagliasacchi2015articulated}. For a position trajectory $p_t$, the first- and second-order residuals are $p_t-p_{t-1}$ and $p_{t+1}-2p_t+p_{t-1}$. We use both for wrist and object translation. For rotations, let $U_t=R_t R_{t-1}^{\top}$ be the increment between adjacent frames. We penalize $\|\log U_t\|_2^2$ for wrists, fingers, and objects. The object term also penalizes $\|\ell(U_{t+1},U_t)\|_2^2$. Object differences are evaluated in world coordinates. For a hand rotation, the weight is multiplied by $1+4(1-v)^2$, where $v$ is the mean visibility over the two frames for that joint (averaged over joints for the wrist).

\paragraph{Hand self-collision ($\mathcal L_{\mathrm{self}}$).} For two capsules on different fingers with radii $r_1,r_2$ and shortest axis distance $d$, the penalty is $\max(0,r_1+r_2-d)^2$, with zero clearance margin.

\paragraph{Object--object nonpenetration ($\mathcal L_{\mathrm{OO}}$).} Let $s$ be the object's signed distance (positive outside) plus the penetration tolerance $\tau$. For smoothing half-width $\epsilon>0$, define $u=(\epsilon-s)/(2\epsilon)$. The residual used before squaring is
\begin{equation}
\phi_\epsilon(s)=
\begin{cases}
-s, & s\leq-\epsilon,\\
\epsilon(2u^3-u^4), & -\epsilon<s<\epsilon,\\
0, & s\geq\epsilon .
\end{cases}
\end{equation} For $\epsilon=0$, we use $\max(0,-s)$.

\paragraph{Hand--object nonpenetration ($\mathcal L_{\mathrm{HO}}$).} When a finger capsule intersects an object but its bone axis does not, we probe the object SDF at both bone endpoints and their midpoint. We use the probe closest to the candidate contact along the axis. Where the signed distance magnitude exceeds $0.1$\,mm, the candidate normal must agree with the SDF gradient (dot product above $0.25$). An exterior probe also requires the face to point toward the axis within a $0.1$\,mm tolerance.

Switching the stored plane to the opposite side requires three consecutive frames of supporting evidence. We release the plane after three consecutive frames with clearance above 12\,mm.

\paragraph{Solver weights.} Table~\ref{tab:supp_motion_weights} gives the coefficients applied to squared residuals. The object visual weight multiplies the sum of the depth term and the silhouette term, with an additional factor of $2$ on the latter. The table gives temporal weights before the visibility adjustment.

\begin{table}[t]
\centering\small
\caption{\textbf{Joint reconstruction weights.} Paired values follow the order in the left column. Translation and depth are in mm, rotation in radians, and silhouette distance in pixels.}
\label{tab:supp_motion_weights}
\begin{tabular}{@{}ll@{}}\hline
Residual or block & Coefficient \\ \hline
Hand depth / object visual block & $1\;/\;0.3$ \\
Hand rotation prior: wrist / finger & $10^6\;/\;10^3$ \\
Wrist translation: first / second order & $150\;/\;150$ \\
Wrist / finger rotation: first order & $2{\times}10^7\;/\;2{\times}10^7$ \\
Object translation: first / second order & $30\;/\;100$ \\
Object rotation: first / second order & $10^3\;/\;10^5$ \\
Hand self-collision & $10^3$ \\
Hand--object / object--object collision & $10^5\;/\;10^7$ \\ \hline
\end{tabular}
\end{table}
The hand--object and object--object penalties both use zero tolerance, with smoothing half-widths of $1$ and $3$\,mm, respectively. Per hand--object pair we use 2,048 object surface samples, 32 samples per finger capsule and 256 palm samples.

\subsubsection{Window schedule and pose updates}

Each window contains up to 121 frames and advances by 90 frames, with the final window shifted to end at the last frame. We take up to 12 Gauss--Newton steps per window with a step scale of $0.5$ and fix the object poses at sequence frame zero. At iteration 4, we activate both nonpenetration terms and compute the pair confidences below. These stay fixed for the remaining iterations of that window.

\subsubsection{Confidence weighting of the contact update}

For hand $h$ and object $o$ at frame $t$, let $N_h$ and $N_o$ count their respective accepted ICP matches. The numbers of matches per pose variable are $s_o=N_o/6$ and $s_h=N_h/28$. The raw object confidence is
\begin{equation}
 c_{ho,t}=\frac{s_o^2}{s_o^2+s_h^2}.
\end{equation}
When both counts vanish, we assign the raw confidence $c_{ho,t}=1$. We apply a five-frame median filter per pair, replicating the boundary values, to obtain $\widetilde c_{ho,t}$. The weight on the coupled update is $\beta_{ho,t}=\lambda_{\rm int}(1-\widetilde c_{ho,t})$, with $\lambda_{\rm int}=0.1$.

Let $Q$ denote the linearized contact cost for a pair and $\Delta o_{\rm ref}$ the object step from the reference solve before adding its hand--object contact terms. For hand and object increments $\Delta h,\Delta o$, the pair contributes
\begin{equation}
 (1-\beta_{ho,t})Q(\Delta h,\Delta o_{\rm ref})+\beta_{ho,t}Q(\Delta h,\Delta o).
\end{equation}
Both terms use the same contact matches. In the DexYCB runs we examined, the median object confidence was $0.976$, giving the coupled term a weight of just $0.24\%$ at the median. Thus, penetration is corrected mainly through the hand pose, while the object update stays close to its visual reference.

\section{Reconstruction Evaluation Protocol}
\label{sec:supp_experiment_protocols}

\paragraph{TACO camera trajectory and evaluation alignment.}
HOForce and HaWoR share an RGB-D SLAM camera trajectory. Dyn-HaMR uses it to initialize camera motion and optimizes the translation scale during fitting. For FoundationPose, ICG, and Point2Pose, we use the trajectory to transform predicted object poses from camera to world coordinates. To align the resulting reconstructions with the annotations, we fit one rigid transform per method and sequence jointly to both object trajectories. For hands, we fit a separate rigid transform to the available wrist positions, giving each hand equal total weight. We apply it to all joints, score each hand on its valid frames, and average the hand scores.

\paragraph{DexYCB initialization.} We use camera \texttt{836212060125}. AGILE shares our hand initialization and uses FoundationPose for its initial object trajectory, with fixed object scale and both depth terms enabled. For HOTrack, we try the annotated hand and object poses at frame zero and advance one frame at a time through frame 20 if initialization fails. Only predictions after the successful start frame are scored.

\paragraph{Pose and motion errors.}
For objects, ADD is the mean distance between corresponding model vertices at the predicted and annotated poses. ADD-S uses the nearest vertex instead. For each metric we report the area under the pose success curve (AUC) over distance thresholds from 0 to 100\,mm, normalized by this range. On DexYCB, we exclude the initialization frame. A missing pose on an annotated evaluation frame counts as a failure at every AUC threshold. Translation and rotation errors use frames with available predictions.

For PA, we fit a similarity transform to the 21 joints of each frame. For WA, we fit one transform across up to 100 consecutive evaluated frames. Windows do not overlap, gaps start a new window, and a final partial window contributes if it contains at least two frames.

Acceleration error uses second differences of object translations or hand joint positions divided by the squared sampling interval. A difference contributes only when all three samples are available at the evaluation stride.

\paragraph{Penetration metrics.}
For each observed hand--object pair, we transform the hand mesh into the object's local frame using their predicted poses. Penetration depth (PD) is the greatest distance from an interior hand vertex to the object surface, or zero if there are no interior vertices. To compute intersection volume (IV), we close the hand mesh at the wrist and count the solid object voxels whose centers lie inside the hand. Each voxel has 1\,mm edges and contributes $10^{-3}\,\mathrm{cm}^3$. An observed, separated pair has zero overlap. We exclude pairs with either pose missing. For object--object pairs, we take the maximum PD over both directions and compute IV on a 1\,mm grid fixed to the first object.

TACO contains multiple hand--object pairs. For PD, we take the maximum across valid pairs in each frame and average over frames with at least one valid pair. For IV, we first average over the valid frames of each pair, then average those values across pairs in the sequence.

\begin{figure}[t]
\centering
\includegraphics[width=\linewidth]{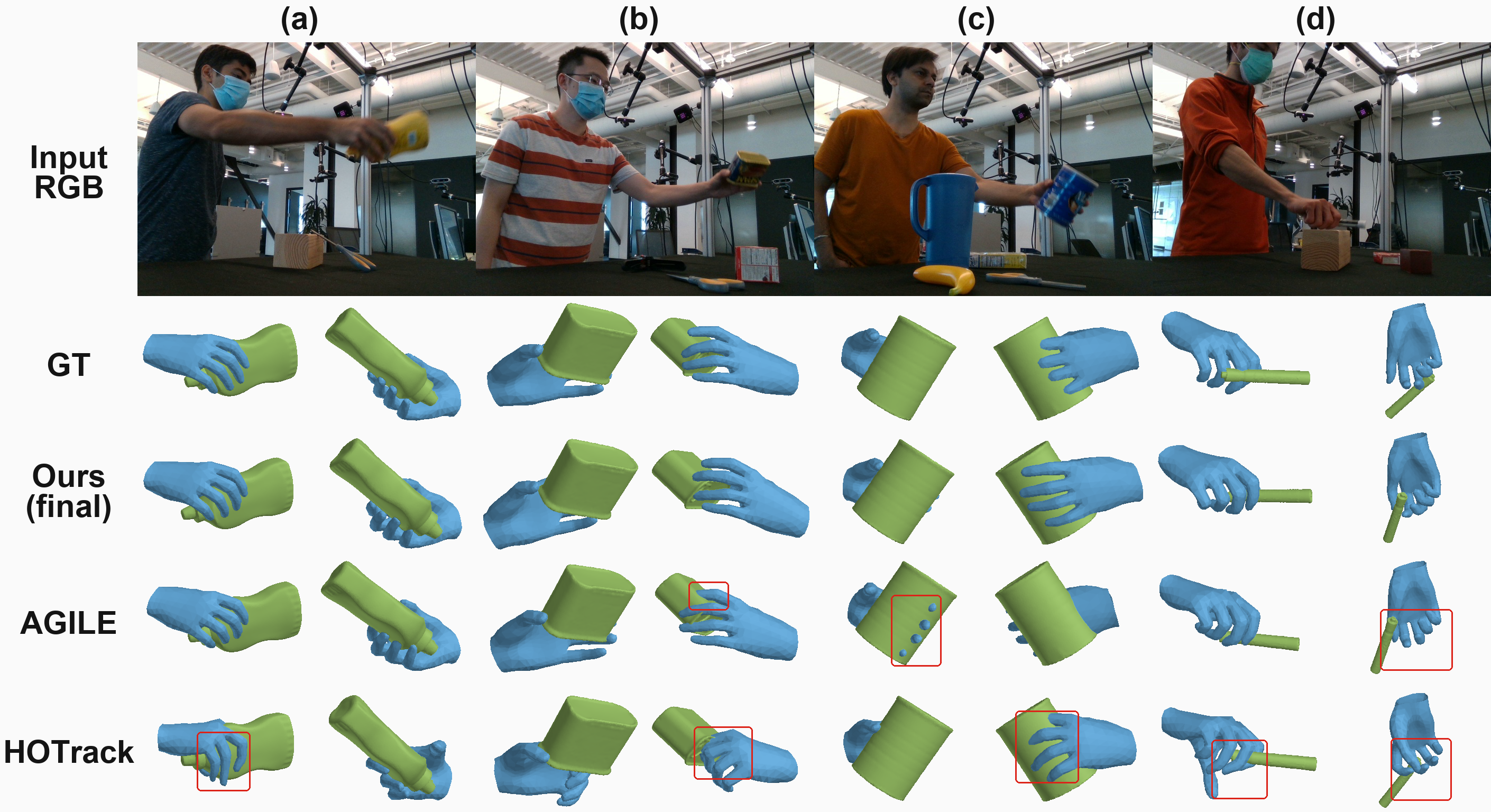}
\caption{Additional DexYCB grasps. Rows show the input RGB, ground truth, our final reconstruction, AGILE, and HOTrack. The two renderings in each cell use the same viewpoints across methods.}
\label{fig:supp_dexycb_grasps}
\end{figure}

\begin{figure}[t]
\centering
\includegraphics[width=\linewidth]{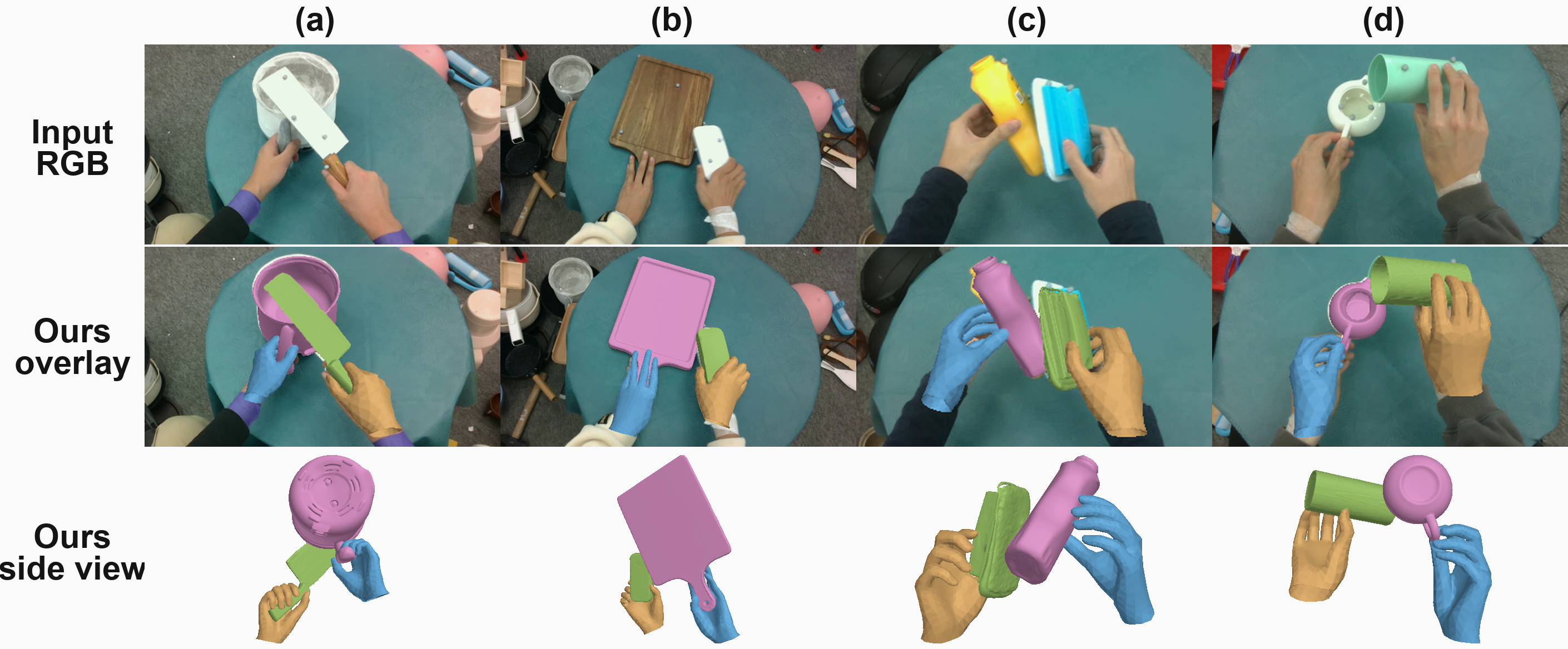}
\caption{Additional bimanual interactions on TACO. Each column shows the input RGB, the reconstruction overlaid on the input, and an alternate view.}
\label{fig:supp_taco_grasps}
\end{figure}

\paragraph{Additional qualitative results.} Figures~\ref{fig:supp_dexycb_grasps} and~\ref{fig:supp_taco_grasps} show four additional examples per dataset.

\section{Force Optimization, Evaluation, and Replay}
\label{sec:supp_force}

\subsection{Pose notation and contact costs}
\label{sec:supp_force_notation}
We use a metric world frame and fixed hand shape. Throughout the main paper and this section, $\widehat T$ denotes the tracked object initialization, $T$ the object trajectory after joint reconstruction, and $\widetilde T$ the trajectory generated by the applied forces. Thus, the reconstructed object geometry used for contact selection is $T$. We interpolate its translation linearly and its rotation on $SO(3)$ between video frames. Visual poses and velocities initialize the first simulation window; subsequent windows inherit the preceding simulated endpoint. All trials within a window start from the same state.

\paragraph{Physical parameters and simulation settings.}
For DexYCB, we use YCB reference masses and compute centers of mass and inertias from the object meshes assuming uniform density. For TACO, we estimate mass, center of mass, and inertia from the mesh with a density of $1000\,\mathrm{kg/m^3}$. For each instrumented object, we use the total mass of the assembled configuration, including its sensors, and compute its center of mass and inertia from the corresponding mesh under uniform density.

Under a 5\,N normal load, Liu and Sung~\cite{liu2002fingertipfriction} measured fingertip kinetic friction coefficients of $1.25\pm0.29$ on grooved polycarbonate and $1.61\pm0.49$ on untextured polycarbonate (mean $\pm$ standard deviation). Guided by these measurements and a large language model's estimate, we set the effective finger--object friction coefficients in Table~\ref{tab:supp_contact_pose_weights}. Each contact cone has $N_r$ rays. We divide each video-frame interval into equal simulation substeps no longer than $h_{\max}$; the table also gives the MuJoCo integration and contact-solver settings.

In the Astra three-dimensional force comparison, both methods use the same masses, centers of mass, inertias, and contact friction. The effective sliding friction coefficients for finger--object, object--object, and object--support contact are 1.5, 1.3, and 1.3, respectively. Cross-engine replay retains the physical parameters and friction cones of the corresponding MuJoCo source scene and adapts only the fingertip forces.

\paragraph{Object pose update and observation weight.}
The object term in Eq.~(9) of the main paper is the tracking prior in Sec.~\ref{sec:supp_motion}.2.1. With translation in meters and rotation in radians, its notation is
\begin{equation}
\begin{aligned}
 w^o_{o,t}&=\gamma_{o,t},\\
 d_o^2(T,\widehat T)&=k_p\|p-\widehat p\|^2
                  +k_R\|\log(R\widehat R^\top)\|^2,
\end{aligned}
\label{eq:supp_object_prior_mapping}
\end{equation}
where $(k_p,k_R)=(9\times10^6,22500)$ and $\gamma_{o,t}$ is the tracking confidence defined there. The factor $10^6$ converts the squared translation residual from meters to millimeters.

For completeness, the reconstruction update of Sec.~\ref{sec:supp_motion}.2.3 has the following form. Let $Q_0$ collect the linearized reconstruction terms other than hand--object contact, $Q_{ho}$ the contact quadratic for a pair, and $\Delta o_{\rm ref}$ the object step from the reference solve before adding that pair's contact terms. We solve
\begin{equation}
\begin{aligned}
 \min_{\Delta\Theta,\Delta o}\quad &Q_0(\Delta\Theta,\Delta o)+\sum_{h,o,t}Q_{ho}^{\beta},\\
 Q_{ho}^{\beta}={}&(1-\beta_{ho,t})Q_{ho}(\Delta\Theta_h,\Delta o_{\rm ref})\\
                 &+\beta_{ho,t}Q_{ho}(\Delta\Theta_h,\Delta o_o),
\end{aligned}
\label{eq:supp_reconstruction_step}
\end{equation}
with $\beta_{ho,t}=\lambda_{\rm int}(1-\widetilde c_{ho,t})$ and $\lambda_{\rm int}=0.1$. The confidence $\widetilde c_{ho,t}$ and the weights within $Q_0,Q_{ho}$ are given in Sec.~\ref{sec:supp_motion}.2 and Table~\ref{tab:supp_motion_weights}. For the resulting object increment $\Delta o=(\Delta p,\Delta\omega)$, the update is
\begin{equation}
 p^+=p+\eta_o\Delta p,\qquad
 R^+=\exp(\eta_o[\Delta\omega]_\times)R,
 \quad\eta_o=0.5.
\label{eq:supp_object_pose_step}
\end{equation}
This defines the reconstructed trajectory $T$ supplied to the force and hand/contact subproblems below as their geometric reference.

\paragraph{Gap, load, and attraction.}
Let $c_i,n_i$ be an object-local surface point and outward unit normal. The force acts at $c_i$; the gap cost uses the target $\bar c_i=c_i-\epsilon_g n_i$, with $\epsilon_g=0.75$\,mm. For the corresponding hand-surface point $v_i(\Theta)$, the two evaluations of the gap are
\begin{equation}
 g_i^T=v_i(\Theta)-T_{o(i)}\bar c_i,\qquad
 g_i^{\rm sim}=v_i(\Theta)-\widetilde T_{o(i)}\bar c_i.
\label{eq:supp_two_gaps}
\end{equation}
They share the same finger--object assignment and hand-mesh correspondence. Define
\begin{equation}
\begin{aligned}
 Q(g,a)&=\frac{w_Qw_n}{2}\left(\frac a{a_0}\right)^2
                       \left\|\frac g{d_0}\right\|^2,\\
 \psi(r)&=\frac12\frac{\zeta^2r^2}{\zeta^2+r^2},\qquad
 \zeta=\delta/d_0.
\end{aligned}
\label{eq:supp_gap_costs}
\end{equation}
Here $a=\mathbf1^\top\alpha/h$ is the ray-load proxy over application duration $h$. For interval $t$ with $S_t$ substeps, define $\langle b\rangle_t=\sum_{s=0}^{S_t}\pi_{t,s}b(s)$, with $\pi_{t,0}=\pi_{t,S_t}=1/(2S_t)$ and $\pi_{t,s}=1/S_t$ at interior nodes. Let $\omega_t=1/K$ for a window of $K$ intervals, and $N_t$ its active-contact count. The accumulated costs are
\begin{equation}
\begin{aligned}
 Q_X&=\sum_{t,i}\langle Q(g_i^X,a_i)\rangle_t,\\
 A_X&=\lambda_{\rm acq}\sum_t\frac{\omega_t}{N_t}
       \left\langle\sum_i\psi\!\left(\frac{\|g_i^X\|}{d_0}\right)\right\rangle_t,
\end{aligned}
\label{eq:supp_gap_aggregation}
\end{equation}
for $X=T$ or ${\rm sim}$. An interval without active contacts contributes zero to both terms. This is the averaging convention in Eq.~(8) of the main paper. Table~\ref{tab:supp_contact_pose_weights} gives all gap coefficients and scales.

\subsection{Contact selection and load initialization}
\label{sec:supp_contact_selection}
At each frame, we assign each reconstructed fingertip to the nearest object and initialize its contact point by projection onto that object's surface. We find the hand-side correspondence within the surface region of the corresponding fingertip. For a layout $P=\{c_i\}$, local search retains the current finger--object assignments and hand-side correspondences within an $R_c$ neighborhood of the current object-surface points. The ray impulses start at zero; candidate layouts provide a load initialization for the subsequent simulation fit.

\paragraph{Motion-derived force and torque.}
For one interval of duration $h$, let $(p_0,R_0,v_0,\omega_0)$ be the initial simulated center-of-mass state and $(p_1,R_1,v_1,\omega_1)$ the endpoint state derived from $T$. Write $\operatorname{cap}_b(x)=x\min(1,b/\|x\|)$, with value zero at $x=0$. To initialize candidate loads, we bias endpoint velocities toward the reference displacement. The resulting velocity seeds and wrench are
\begin{equation}
\begin{aligned}
 v_d&=v_1+\operatorname{cap}_{b_v}\!\left(\frac{p_1-p_0}{2h}\right),\\
 \omega_d&=\omega_1+\operatorname{cap}_{b_\omega}\!\left(\frac{\log(R_1R_0^\top)}{2h}\right),\\
 F^\star&=m\left[(v_d-v_0)/h-\mathbf g\right],\\
 \tau^\star&=(I_1\omega_d-I_0\omega_0)/h.
\end{aligned}
\label{eq:supp_layout_wrench}
\end{equation}
$I_0,I_1$ are world-frame inertias, $\mathbf g$ is gravity, and $(b_v,b_\omega)=(0.5\,\mathrm{m/s},5\,\mathrm{rad/s})$. We stack the wrenches as $w^\star$ to rank layouts. The subsequent MuJoCo fit resolves object--object and support reactions together with the applied fingertip forces.

\paragraph{Candidate score.}
For each contact, we select up to $N_c$ candidate points near the corresponding fingertip from distinct surface regions in the local neighborhood, preferentially retaining points whose normals differ from those of already selected candidates by more than $\theta_c$. Each search changes at most $K_c$ contact points and evaluates up to $N_P$ layouts. The scores below shortlist at most $N_H$ layouts for shared hand-pose fitting.

Let $u_j\geq0$ be the normal load on cone ray $j$ and $d_j$ its direction scaled to unit inward normal component. At the initial orientation, the ray contributes the column $[d_j^\top,((p_i-p_{\rm com})\times d_j)^\top]^\top$ to the wrench matrix $A(P)$. We solve
\begin{equation}
 E_{\rm fit}(P)=\min_{u\in\mathcal U}
 \|S^{-1}(A(P)u-w^\star)\|_1
 +\lambda_u\sum_{j:\bar u_j>0}\frac{u_j}{\bar u_j}.
\label{eq:supp_layout_lp}
\end{equation}
$\mathcal U$ imposes $u\geq0$, $\sum_{j\in(h,f)}u_j\leq B_f$ for each finger $f$, and $\sum_{j\in h}u_j\leq B_h$ for each hand. Here $\bar u_j=\min(B_f,B_h)$. For each object, $S$ scales force by $s_F=\max(m\|\mathbf g\|,0.5\,\mathrm N)$ and torque by $s_F\ell$, where $\ell$ is one quarter of the bounding-box diagonal, with a 10\,mm minimum. If $d_i^{\rm tip}$ is the distance to the corresponding fingertip region and $c_i^0$ is the current point, the layout score is
\begin{equation}
 E_{\rm rank}(P)=E_{\rm fit}(P)
 +\sum_i\frac{\lambda_d(d_i^{\rm tip})^2
                   +\lambda_p\|c_i-c_i^0\|^2}{R_c^2}.
\label{eq:supp_layout_rank}
\end{equation}
We fit the shared hand pose for each shortlisted layout using Sec.~\ref{sec:supp_hand_update}, project its points onto the selected surface patches, and refit Eq.~\eqref{eq:supp_layout_lp}. Let $\mathcal V$ contain candidates with a successful refit and non-worsening finger self-collision, $E_k$ their refitted wrench costs, and $H_k$ their hand costs excluding contact-fit terms. The choice is
\begin{equation}
\begin{aligned}
 E_{\min}&=\min_{k\in\mathcal V}E_k,\\
 \mathcal T&=\{k\in\mathcal V:E_k-E_{\min}\leq
                         \epsilon_a+\epsilon_rE_{\min}\},\\
 k^\star&=\min\!\left(\operatorname*{arg\,min}_{k\in\mathcal T}H_k\right),
 \quad(\epsilon_a,\epsilon_r)=(10^{-8},10^{-3}).
\end{aligned}
\label{eq:supp_layout_acceptance}
\end{equation}
The outer minimum breaks an exact tie by candidate index. If $\mathcal V$ is empty, we retain the current layout. For unit cone rays with inward normal component $\eta_j$, the selected normal loads initialize impulses as $\alpha_j^{(0)}=h u_j/\eta_j$.

\subsection{Hand-pose update and simulation acceptance}
\label{sec:supp_hand_update}
All contacts on one hand share its 51 pose variables: wrist translation and rotation, and 15 joint rotations. For the normal load $\ell_i$ on contact $i$, the hand contact-fit weight is
\begin{equation}
 \omega_i^{\rm IK}=\lambda_{\rm IK}
 \left[\rho_0+(1-\rho_0)\min(\ell_i/a_0,1)\right].
\label{eq:supp_ik_weight}
\end{equation}
With $(\lambda_{\rm IK},\rho_0)=(10^6,0.1)$, even an unloaded contact contributes to the hand fit. For fixed candidate points, the hand objective is
\begin{equation}
\begin{aligned}
 K_h(\Theta_h)={}&\frac12\sum_{i\in h}\omega_i^{\rm IK}\|g_i^T\|^2
                  +H_h(\Theta_h),\\
 H_h={}&E_{\rm obs}^h+E_{\rm time}^h+E_{\rm skin}^h
        +E_{\rm prior}^h+E_{\rm anat}^h+E_{\rm self}^h.
\end{aligned}
\label{eq:supp_hand_objective}
\end{equation}

\paragraph{Hand observation and pose prior.}
For joint confidence $c_j\in[0,1]$, the hand term in Eq.~(9) of the main paper uses
\begin{equation}
\begin{aligned}
 z_j&=\operatorname{clip}((c_j-c_0)/(1-c_0),0,1),\\
 w_j^h&=c_j[1+(\kappa_v-1)z_j^2(3-2z_j)],\\
 \Sigma_j^{-1}&=(b_j/\sigma_j^2)I_3,\qquad
 E_{\rm obs}^h=\frac12\sum_jw_j^h
       \|J_j(\Theta_h)-\widehat J_j\|^2_{\Sigma_j^{-1}}.
\end{aligned}
\label{eq:supp_hand_observation}
\end{equation}
$b_j$ is 1 at the wrist and $(0.95,0.75,0.55,0.35)$ along each finger from its first joint to the tip. The distal joint and tip use $\sigma_j=8$\,mm; the other joints use 5\,mm. We use $c_0=0.75$ and $\kappa_v=1000$.

Let $q=\sum_jb_jc_j/\sum_jb_j$ and let $q_j$ be the maximum confidence from finger joint $j$ to its tip. The wrist prior has standard deviations $\sigma_p(q)=(1-q)\sigma_p^-+q\sigma_p^+$ and $\sigma_R(q)=(1-q)\sigma_R^-+q\sigma_R^+$. Relative to the visual hand pose, it contributes
\begin{equation}
\begin{aligned}
 E_{\rm prior}^h=\frac12\bigg[&\frac{\|p_h-\widehat p_h\|^2}{\sigma_p(q)^2}
 +\frac{\|\log(R_h\widehat R_h^\top)\|^2}{\sigma_R(q)^2}\\
 &+\sum_{j=1}^{15}b_j^\theta(0.05+0.95q_j)
                    \|\log(R_j\widehat R_j^\top)\|^2\bigg],
\end{aligned}
\label{eq:supp_hand_prior}
\end{equation}
where $b_j^\theta=(1,0.9,0.7)$ from proximal to distal. Table~\ref{tab:supp_contact_pose_weights} gives the wrist scales. The anatomical cost is $E_{\rm anat}^h=\sum_j\|a_j-\operatorname{clip}(a_j,a_j^-,a_j^+)\|^2$, with $a_j$ the joint rotation components in its anatomical axes and $[a_j^-,a_j^+]$ the hand model's allowed interval. For capsule overlap $d_k$, $E_{\rm self}^h=\tfrac12\sum_k([d_k-\epsilon_c]_+/\sigma_c)^2$, where $[x]_+=\max(x,0)$ and $\epsilon_c=\sigma_c=2$\,mm. The reconstruction priors preceding this update are specified in Sec.~\ref{sec:supp_motion}.2.1 and Table~\ref{tab:supp_motion_weights}.

\paragraph{Temporal and surface continuity.}
Let $h_t=\tau_t-\tau_{t-1}$, $\bar h_t=(h_t+h_{t-1})/2$, $h_*=1/30$\,s, and $D_ty=(y_t-y_{t-1})/h_t$. For the 21 hand joints, set $d_j=D_tJ_j-D_{t-1}J_j$. Let $e_R$ be the wrist rotation error relative to constant-angular-velocity continuation from its two preceding poses. Then
\begin{equation}
 E_{\rm time}^h=\frac{w_h\kappa_h h_*^3}{2\bar h_t}
 \left[\frac{\sum_j\|d_j\|^2}{21\sigma_J^2}
      +\frac{\|e_R\|^2}{h_t^2\sigma_\omega^2}\right].
\label{eq:supp_hand_temporal}
\end{equation}
For a persistent hand-mesh contact, let $\Delta p_s,\Delta n_s$ be the changes of its point and unit normal in the reconstructed object's coordinates, and $L$ the smaller of the finger's peak force magnitudes in the two intervals. Its contribution to $E_{\rm skin}^h$ is
\begin{equation}
 \frac{w_s}{2}\frac{L}{L+a_0}\left(\frac{h_*}{h_t}\right)^2
 \left[\frac{\|\Delta p_s\|^2}{s_c^2}
      +\frac{\|\Delta n_s\|^2}{s_n^2}\right].
\label{eq:supp_skin_continuity}
\end{equation}
Terms requiring history enter once that history is available.

\paragraph{Pose increment and acceptance.}
Stack the weighted residuals of Eq.~\eqref{eq:supp_hand_objective} as $r_h$, with Jacobian $J_h$. In scaled coordinates $\Delta\theta_h=D_hz$, we compute
\begin{equation}
 z^\star=\operatorname*{arg\,min}_z
 \frac12\|r_h+J_hD_hz\|^2+\frac{\lambda_{\rm GN}}2\|z\|^2,
 \qquad\lambda_{\rm GN}=10^{-8}.
\label{eq:supp_hand_gn}
\end{equation}
$D_h$ uses scales of 2\,mm for wrist translation, $3^\circ$ for wrist rotation, and $10^\circ$ for finger rotations. We rescale the direction to satisfy $\|z\|\leq1$ and a 3\,mm maximum linearized joint/contact-point displacement, and project it against the linearized capsule-overlap constraints. For step sizes $\eta\in\{1,1/2,1/4\}$, translation uses $p_h^+=p_h+\eta\Delta p_h$ and each rotation uses $R_j^+=\exp(\eta[\Delta\omega_j]_\times)R_j$. We recompute the hand geometry and accept the first step satisfying
\begin{equation}
\begin{aligned}
 K_h(\Theta_h^+)&<K_h(\Theta_h)-\epsilon_K,\\
 [d_k(\Theta_h^+)-\epsilon_c]_+
   &\leq[d_k(\Theta_h^0)-\epsilon_c]_++\epsilon_{\rm col}
      \quad\forall k,
\end{aligned}
\label{eq:supp_hand_acceptance}
\end{equation}
where $k$ indexes capsule pairs, $\Theta_h^0$ is the pose before the hand solve, $\epsilon_K=10^{-10}$, and $\epsilon_{\rm col}=10^{-6}$\,m. If no step passes, the hand pose is retained.

\paragraph{Force refit and complete candidate score.}
At each proposed geometry we refit the cone impulses by linearizing motion/contact residuals, solving damped least squares, and projecting onto nonnegativity and load limits. Each projected force trial is evaluated by MuJoCo from the window's initial state. A force update must not increase the motion error or physical cost. The normalized motion residual is
\begin{equation}
 r_o=\left[\frac{\Delta p}{s_p},\frac{\log(\widetilde RR^\top)}{s_R},
           \frac{\Delta v}{s_v},\frac{\Delta\omega}{s_\omega}\right],
\label{eq:supp_dynamics_scales}
\end{equation}
where $\Delta$ compares the simulated state to the state derived from $T$. Position residuals use object-frame origins, and linear velocities refer to the center of mass. We append object-pair relative translation and rotation residuals with the same pose scales. The force-fit objective is
\begin{equation}
 E_{\rm phys}=\frac{w_{\rm dyn}}2\sum_{t,o}\omega_t\|r_{o,t}\|^2
              +Q_{\rm sim}+A_{\rm sim}+E_F+E_b+E_C.
\label{eq:supp_physical_score}
\end{equation}
The last three terms regularize the force history, interval boundaries, and object-surface contact continuity. Using the time differences defined above, let $f_t$ be the world-frame force at the first application of interval $t$, for one finger--object pair. Its force-history contribution is
\begin{equation}
 \frac{w_{F1}h_*}{2h_t}\|f_t-f_{t-1}\|^2
 +\frac{w_{F2}h_*^3}{2\bar h_t}\|D_tf-D_{t-1}f\|^2.
\label{eq:supp_force_temporal}
\end{equation}
Summing over intervals and pairs gives $E_F$. At an interval boundary, let $f^-,f^{--}$ be the last two previously applied forces, $\Delta s_-$ their time separation, and $\Delta s_+$ the time from $f^-$ to the new first force $f_t$. The boundary contribution is
\begin{equation}
 \frac{\kappa_b w_{F1}h_*}{2h_t}
 \left[\|f_t-f^-\|^2+
 \left\|f_t-f^--\frac{\Delta s_+}{\Delta s_-}(f^--f^{--})\right\|^2\right].
\label{eq:supp_force_boundary}
\end{equation}
Their sum is $E_b$. For these differences, a missing finger--object force is zero.

For object-local contact continuity, set $q_a=\min(1,a^-/a_0)$ using the preceding normal load. Relative to the preceding interval's last application point and unit normal, a current point and normal contribute
\begin{equation}
 w_Cq_a\left[\rho\!\left(\frac{\Delta c}{v_C\Delta t_C}\right)
       +\frac12\left\|\frac{\Delta n}{\nu_C\Delta t_C}\right\|^2\right],
\label{eq:supp_object_contact_temporal}
\end{equation}
Here $\rho(x)=\sqrt{1+\|x\|^2}-1$. We average this contribution at the first and last applications of the current interval, using the elapsed time $\Delta t_C$ between the last applications. Ending a loaded contact or switching its object contributes $w_Cq_a$. Summing these terms gives $E_C$.

To compare complete hand/contact candidates, we replace the simulated-gap terms by their reconstructed-gap counterparts and add the hand observation and continuity costs:
\begin{equation}
\begin{aligned}
 E_{\rm sel}={}&E_{\rm phys}-Q_{\rm sim}-A_{\rm sim}+Q_T+A_T\\
               &+\sum_h(E_{\rm obs}^h+E_{\rm time}^h+E_{\rm skin}^h).
\end{aligned}
\label{eq:supp_complete_selection}
\end{equation}
The hand-pose prior, anatomical, and self-collision costs enter $K_h$ in Eq.~\eqref{eq:supp_hand_objective}; the final comparison uses Eq.~\eqref{eq:supp_complete_selection} with the geometric acceptance checks above. The current candidate participates in the comparison, and the lowest-scoring valid candidate is retained. We first optimize individual frame intervals, then refine windows of up to five adjacent intervals, with 2--6 iterations per stage.

The block weights in Eq.~(6) of the main paper are $\lambda_{\rm dyn}=w_{\rm dyn}$ and $\lambda_{\rm contact}=w_Q$, with $\lambda_{\rm obs}=\lambda_{\rm prior}=\lambda_{\rm temp}=1$; the observation, prior, and temporal blocks include their per-term coefficients. The contact-fit coefficient $\lambda_{\rm IK}$ belongs to the hand subproblem in Eq.~\eqref{eq:supp_hand_objective}.

\begin{table*}[t]
\centering\small
\caption{\textbf{Force optimization and physical parameters.} Symbols are defined in the text and equations above. Coefficients multiply costs, rather than residuals. All geometric residuals below use meters; angular residuals use radians. Reconstruction parameters inherited from Sec.~\ref{sec:supp_motion} are listed separately from the force/contact updates.}
\label{tab:supp_contact_pose_weights}
\begin{tabular}{@{}lll@{}}
\hline
Term or setting & Symbol & Value \\ \hline
Main Eq.~(6): dynamics / contact & $\lambda_{\rm dyn},\ \lambda_{\rm contact}$ & $10^6,\ 1$\\
Main Eq.~(6): observation / prior / temporal & $\lambda_{\rm obs},\ \lambda_{\rm prior},\ \lambda_{\rm temp}$ & $1,\ 1,\ 1$\\
\hline
\multicolumn{3}{@{}l}{\textit{Reconstructed object trajectory (Sec.~\ref{sec:supp_motion}.2)}}\\
Object observation metric: translation / rotation & $k_p,\ k_R$ & $9\times10^6,\ 22500$\\
Contact coupling / object step scale & $\lambda_{\rm int},\ \eta_o$ & $0.1,\ 0.5$\\
\hline
\multicolumn{3}{@{}l}{\textit{Contact selection and force fitting}}\\
Ray-load cost & $\lambda_u$ & $10^{-4}$\\
Fingertip distance / point shift & $\lambda_d,\ \lambda_p$ & $0.002,\ 0.002$\\
Surface search radius & $R_c$ & $45$\,mm\\
Maximum candidate points per contact & $N_c$ & $3$\\
Preferred candidate-normal separation & $\theta_c$ & $45^\circ$\\
Maximum changed contact points per search & $K_c$ & $2$\\
Maximum evaluated layouts per search & $N_P$ & $64$\\
Maximum layouts for shared hand-pose fitting & $N_H$ & $2$\\
Load--gap outer / node coefficient & $w_Q,\ w_n$ & $1,\ 10$\\
Attraction weight / robust scale & $\lambda_{\rm acq},\ \delta$ & $10,\ 5$\,mm\\
Load / gap scale & $a_0,\ d_0$ & $1$\,N, $1$\,mm\\
Gap-target inward offset & $\epsilon_g$ & $0.75$\,mm\\
Dynamics tracking & $w_{\rm dyn}$ & $10^6$\\
Position / rotation scale & $s_p,\ s_R$ & $0.01$\,m, $0.1$\,rad\\
Linear / angular velocity scale & $s_v,\ s_\omega$ & $0.3$\,m/s, $3$\,rad/s\\
Finger caps, thumb to little / total per hand & $B_f,\ B_h$ & $(30,25,25,20,15)$\,N / $60$\,N\\
Force first / second differences & $w_{F1},\ w_{F2}$ & $1920,\ 1920$\\
Force boundary factor & $\kappa_b$ & $0.25$\\
Object-surface contact continuity & $w_C$ & $120$\\
Contact point / normal rates & $v_C,\ \nu_C$ & $0.15$\,m/s, $2.5\pi$\,s$^{-1}$\\
\hline
\multicolumn{3}{@{}l}{\textit{Hand-pose fitting}}\\
Contact-fit coefficient / unloaded fraction & $\lambda_{\rm IK},\ \rho_0$ & $10^6,\ 0.1$\\
Visibility transition / maximum multiplier & $c_0,\ \kappa_v$ & $0.75,\ 1000$\\
Proximal / distal observation scale & $\sigma_j$ & $5$\,mm / $8$\,mm\\
Wrist translation prior: low / high confidence & $\sigma_p^-,\ \sigma_p^+$ & $50$\,mm / $6$\,mm\\
Wrist rotation prior: low / high confidence & $\sigma_R^-,\ \sigma_R^+$ & $60^\circ$ / $15^\circ$\\
Hand temporal coefficient / strength & $w_h,\ \kappa_h$ & $10^4,\ 0.25$\\
Temporal joint / wrist-rotation scale & $\sigma_J,\ \sigma_\omega$ & $5$\,mm, $15^\circ$\\
Hand-surface continuity & $w_s$ & $4$\\
Hand-surface point / normal scale & $s_c,\ s_n$ & $1.5$\,mm, $0.1$\\
Capsule allowance / residual scale & $\epsilon_c,\ \sigma_c$ & $2$\,mm, $2$\,mm\\
Scaled Gauss--Newton damping & $\lambda_{\rm GN}$ & $10^{-8}$\\
\hline
\multicolumn{3}{@{}l}{\textit{Physical parameters and MuJoCo settings for our method}}\\
Finger--object friction: DexYCB and TACO & $\mu$ & $1.3$\\
Finger--object friction: 32-sequence force evaluation & $\mu$ & $1.5$\\
Friction-cone rays per contact & $N_r$ & $16$\\
Maximum simulation substep & $h_{\max}$ & $2$\,ms\\
MuJoCo integrator & --- & \texttt{implicitfast}\\
MuJoCo contact solver & --- & Newton\\
Contact-solver iteration limit / tolerance & --- & $100$ / $10^{-10}$\\
Contact response & \texttt{solref} & $(0.004,1)$\\
Contact impedance & \texttt{solimp} & $(0.95,0.99,0.001,0.5,2)$\\
Collision margin / gap & --- & $0$ / $0$\\
\hline
\end{tabular}
\end{table*}

\subsection{Force evaluation and replay}

\begin{figure*}[t]
\centering
\includegraphics[width=\textwidth]{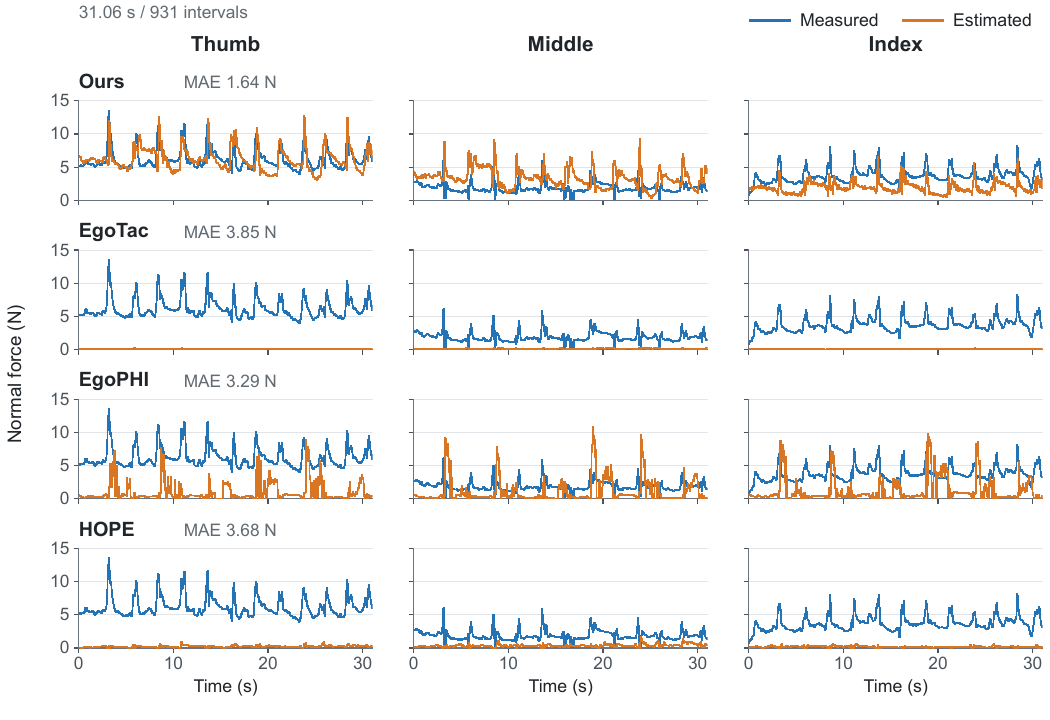}
\caption{\textbf{Force reconstruction for all three measured fingers.} Rows compare each method with the measured normal force; columns show the thumb, middle, and index fingers. All panels use identical axes and the same 931 frame intervals over the complete 31-second sequence. Row labels report MAE across all three fingers.}
\label{fig:supp_force_curves_all}
\end{figure*}

\paragraph{Measured-force protocol.}
We align sensor samples to video frames by their host timestamps, rotate the measured forces into world coordinates, and average them over each frame interval. For each interval, we pair the force estimated by HOForce with the reading from the sensor whose active area contains the contact point. In all valid evaluation intervals, HOForce's predicted contact points lie within the active areas of their corresponding sensors. No samples are excluded because a predicted contact point lies outside the sensor's active area. For the normal-force comparison in the main paper, we take the absolute projection of each vector onto the same sensor normal.

We compare normal forces across the three open-source baselines because this quantity can be read from all three outputs. For EgoTac~\cite{zhang2026egotac}, we map the sensor center to the closest triangle on the hand mesh and interpolate its three nonnegative vertex forces using barycentric coordinates. For EgoPHI~\cite{ilic2026egophi}, we multiply each object-surface force vector by its predicted contact probability, average these vectors over the sensor face using area weights, and take the absolute projection onto the sensor normal. For HOPE~\cite{jeon2026hope}, we integrate the nonnegative part of its released contact-weighted pressure over the hand surface facing the sensor, within its footprint and within 10\,mm of its plane. We report the compressive component along the sensor normal.

We compare alternative force readouts and select a well-performing option for each open-source method. Each method uses the selected readout consistently across all evaluation sequences.

For sequence $i$ with $N_i$ common valid sensor--interval samples, let $a_i=N_i^{-1}\sum_j|\hat f_{ij}-f_{ij}|$ and $e_i=N_i^{-1}\sum_j(\hat f_{ij}-f_{ij})^2$. We report $\mathrm{MAE}=S^{-1}\sum_i a_i$ and $\mathrm{RMSE}=\sqrt{S^{-1}\sum_i e_i}$. Thus each sequence has equal total weight regardless of duration or sensor count. Zero- and small-force samples are included.

\begin{figure*}[t]
\centering
\includegraphics[width=\textwidth]{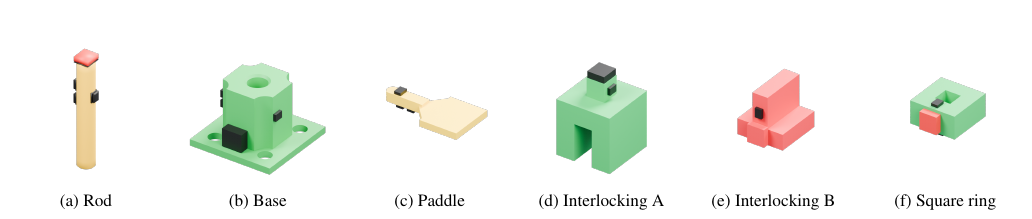}
\caption{\textbf{Object models used in the measured-force sequences.}}
\label{fig:supp_measured_objects}
\end{figure*}

\paragraph{Measured-force sequences.}
We evaluate all methods on the same 32 sequences, spanning six object geometries (Fig.~\ref{fig:supp_measured_objects}) and seven interaction settings. The settings include single-object manipulation of a rod, base, paddle, interlocking part, or square ring, as well as rod--hole insertion and the assembly or disassembly of two interlocking parts. HOForce achieves lower normal-force MAE than HOPE on 31 of the 32 sequences. Figure~\ref{fig:supp_force_curves_all} shows all three measured fingers for the sequence whose thumb trace appears in the main paper.

\paragraph{Data collection and consent.}
The measured-force sequences were recorded by the authors performing everyday object-manipulation tasks; no other participants were recruited. All participating authors consented to the use of their recordings for research and presentation in this paper and its supplementary material. No formal ethics review was sought because data collection was limited to these self-demonstrations by the authors.

\paragraph{Three-dimensional force evaluation.}
We evaluate the full force vectors on all 32 measured-force sequences and compare HOForce with a normal-only version of the same predictions. For the unit sensor normal $\mathbf n$, the normal-only prediction is $\hat{\mathbf f}_{\rm n}=(\mathbf n^\top\hat{\mathbf f})\mathbf n$; its tangential component is zero. This is a post-processing comparison: the predicted normal component and valid sensor--interval samples are unchanged, and no force fitting is repeated. Both predictions are evaluated against the same measured force vectors. The 3D error is the Euclidean distance between predicted and measured vectors; the tangential error is the Euclidean distance between their projections onto the sensor's tangent plane. Table~\ref{tab:supp_tangential_force} reports mean 3D error, mean tangential error, and 3D RMSE. Unlike the normal-force comparison in the main paper, the 3D metrics evaluate the full vector in both rows.

\begin{table}[t]
\centering
\small
\setlength{\tabcolsep}{4pt}
\caption{\textbf{Effect of tangential force estimates on all 32 sequences.} Both rows use HOForce predictions. Normal only retains the predicted normal component and sets the tangential component to zero. All errors are in N; lower is better.}
\label{tab:supp_tangential_force}
\begin{tabular}{@{}lccc@{}}
\hline
Prediction & \shortstack{Mean 3D\\error} & \shortstack{Mean tangential\\error} & \shortstack{3D\\RMSE}\\
\hline
Normal only & 1.4858 & 0.8887 & 2.1016\\
Full vector & \textbf{1.2216} & \textbf{0.5248} & \textbf{1.6892}\\
\hline
\end{tabular}
\end{table}

\paragraph{Comparison with the Astra agent.}
The eight sequences used for the Astra comparison were randomly selected from the 32 measured-force sequences. The Astra agent builds a MuJoCo simulation from RGB-D images, masks, and meshes and refines contacts using visual observations and simulation feedback. Both methods use the same object masses, inertias, and friction coefficients. Each sensor corresponds to a particular hand, finger, and object. For Astra, we sum all forces that finger applies to that object at each simulation step, regardless of the distance from the contact point to the sensor. A step with no such contact contributes zero. We average the resulting force over each video-frame interval, weighted by step duration. For each sensor and frame interval valid for both methods, we compute the Euclidean distance between estimated and measured force vectors in a common world frame and give the eight sequence means equal weight.

\paragraph{Cross-engine insertion.}
In PyBullet and Isaac Sim, we start each video-frame interval with the MuJoCo forces at the original application points. We simulate the interval in the target engine and use the endpoint errors in object pose, linear and angular velocity, and rod--base relative pose to update the three-dimensional forces by damped least squares. We project candidate forces onto the original friction cones and enforce the force limits, with penalties for deviations from the source forces and for changes across intervals.

The first interval starts from the source initial state; each subsequent interval inherits the preceding accepted terminal state in the target engine. All candidate forces within an interval are evaluated from this same initial state. Application points and forces are stored in object-local coordinates. At each simulation substep, a point and force applied to object $o$ are transformed using its current pose $(R_o^E,\mathbf x_o^E)$ in the target engine:
\begin{equation}
\mathbf p_i^W=\mathbf x_o^E+R_o^E\mathbf p_i^B,
\qquad
\mathbf f_i^W=R_o^E\mathbf f_i^B.
\label{eq:supp_target_engine_transport}
\end{equation}
Here $B$ and $W$ denote the object and world coordinate frames, and $E$ denotes the target engine. After adaptation, we replay the complete force sequence continuously from the initial state.

\begin{figure*}[t]
\centering
\includegraphics[width=\textwidth]{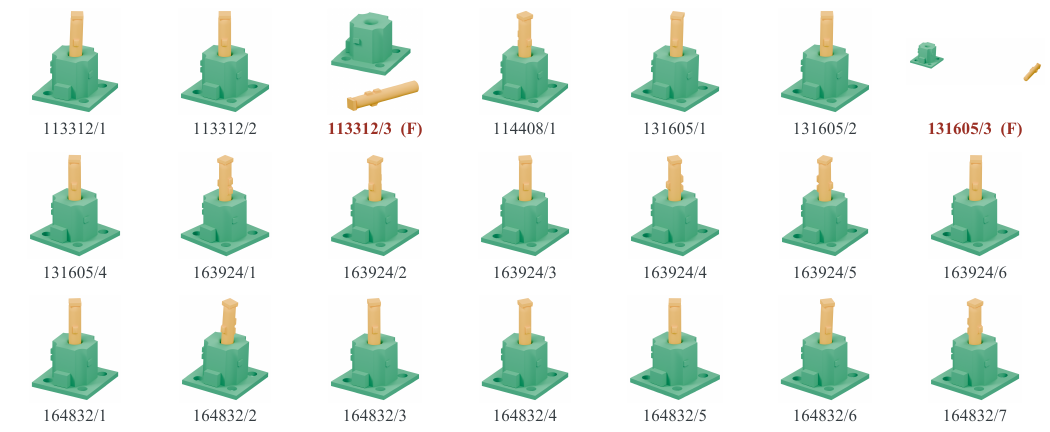}
\caption{Final states of all 21 PyBullet replays after force adaptation. The rod is inserted in 19 sequences. The two failures are marked (F). Labels identify the recording and segment; each view is framed to show both objects.}
\label{fig:supp_pybullet_endframes}
\end{figure*}

Insertion success is determined from the final rod and bore geometry. In the bore's local frame, the rod must enter through the opening, reach at least half the effective bore depth, and pass the wall and bottom geometry checks with a 2\,mm penetration tolerance. We use the same criterion for cross-engine evaluation and the object--object contact ablation. Direct transfer succeeds in 0/21 sequences in both PyBullet and Isaac Sim. After force adaptation, each engine succeeds in 19/21, with 18 successes in common. Figure~\ref{fig:supp_pybullet_endframes} shows the final frame of every PyBullet replay.

\paragraph{Object--object contact ablation.}
We use the same full-method MuJoCo result as in the cross-engine experiment. For the ablation, we exclude contact between the rod and the base only during force fitting. Both force sequences are replayed continuously from the initial state in MuJoCo, with rod--base contact response enabled in both cases. The fitted forces are kept unchanged during replay, and insertion is scored at the final frame. Normal-force MAE/RMSE change from 1.020/1.597~N to 1.021/1.603~N, while valid insertions decrease from 20/21 to 0/21.

\paragraph{Full-set continuous replay.}
The main paper reports replay pose errors on 177 DexYCB videos and 228 TACO videos. Each replay starts from its initial state and applies the saved contact forces continuously, with MuJoCo resolving object--object and support contacts. Replayed positions and velocities match the saved simulated states at every substep, and the full integration state matches at every video-frame boundary. Across the full replay archive, the median per-video translation and rotation RMSEs relative to the visual trajectories used to fit the forces are 3.49\,mm and $2.25^\circ$ on DexYCB and 9.47\,mm and $4.66^\circ$ on TACO.

\section{Additional Experiments}
\subsection{Penetration Ablations}
We ablate the hand--object and object--object nonpenetration losses to measure their effect on mesh overlap across 228 TACO sequences and 21 rod--hole insertion segments from five recordings. Removing nonpenetration penalties increases both penetration depth and intersection volume on both datasets (Tables~\ref{tab:taco_penetration_ablation} and~\ref{tab:insertion_penetration}).

TACO contains many thin-walled objects. When a finger penetrates a thin wall, the nearest surface can be the opposite face, so a standard signed-distance penalty can push the finger farther through the wall. We therefore compare this penalty with our skeleton-guided hand--object correction across all 228 TACO sequences, using the same surface samples and fixed object tracks. Skeleton guidance reduces hand--object PD from 3.38 to 1.73\,mm and IV from 0.64 to 0.20\,cm$^3$ relative to the bidirectional SDF (Table~\ref{tab:taco_penetration_ablation}).

\begin{table}[ht]
\centering
\caption{Penetration on 228 TACO sequences. The hand variants remove the hand--object loss or replace its correction with a bidirectional signed-distance penalty.}
\label{tab:taco_penetration_ablation}
\small
\setlength{\tabcolsep}{5.5pt}
\begin{tabular}{llrr}
\toprule
Pair & Stage~1 variant & PD (mm)$\downarrow$ & IV (cm$^3$)$\downarrow$\\
\midrule
Object--object & w/o $\mathcal L_{\mathrm{OO}}$ & 0.31 & 0.15\\
 & Full & \textbf{0.12} & \textbf{0.02}\\
\midrule
Hand--object & w/o $\mathcal L_{\mathrm{HO}}$ & 3.81 & 0.90\\
 & Bidirectional SDF & 3.38 & 0.64\\
 & Full & \textbf{1.73} & \textbf{0.20}\\
\bottomrule
\end{tabular}
\end{table}

\begin{table}[ht]
\centering
\caption{Penetration on 21 recorded rod--hole insertion segments. Both mean reference IVs round to 0.00\,cm$^3$.}
\label{tab:insertion_penetration}
\small
\begin{tabular}{llrr}
\toprule
Pair & Reconstruction & PD (mm)$\downarrow$ & IV (cm$^3$)$\downarrow$\\
\midrule
Rod--base & Both losses off & 2.60 & 7.21\\
 & Reference tracks & \textbf{0.06} & \textbf{0.00}\\
\midrule
Hand--object & Both losses off & 5.50 & 0.71\\
 & Reference tracks & \textbf{0.15} & \textbf{0.00}\\
\bottomrule
\end{tabular}
\end{table}

\subsection{Fingertip Contact Recovery on DexYCB}
To test whether physical refinement improves fingertip contact recovery, we compare reconstructions before and after refinement on the same 177 DexYCB sequences. For both annotated and reconstructed poses, we mark contact if any fingertip vertex lies within the chosen distance of an object mesh vertex. We average scores equally over sequences. At 3\,mm, physical refinement increases recall from 63.19\% to 68.42\% and F1 from 67.63\% to 70.98\%. At 1\,mm, recall rises from 35.65\% to 44.86\% and F1 from 40.86\% to 47.75\%.

\end{document}